%% file: main.tex
\pdfoutput=1

\documentclass[11pt]{article}

\usepackage[preprint]{acl}

\usepackage{times}
\usepackage{latexsym}

\usepackage[T1]{fontenc}

\usepackage{enumitem}
\usepackage{amsmath}
\usepackage{graphicx}
\usepackage{booktabs}
\usepackage{longtable}
\usepackage{lscape}
\usepackage[utf8]{inputenc}

\usepackage{microtype}

\usepackage{inconsolata}

\title{
Dynamic Multi-Reward Weighting for Multi-Style Controllable Generation
}

\author{Karin de Langis \and Ryan Koo \and Dongyeop Kang \\
University of Minnesota \\
\texttt{\{dento019, koo00017, dongyeop\}@umn.edu}}

\begin{document}
\maketitle
\begin{abstract} 
Textual style expresses a diverse set of information, including interpersonal dynamics (e.g., formality) and the author’s emotions or attitudes (e.g., disgust). An open question is how language models can be explicitly controlled so that they weave together target styles when generating text: for example, to produce text that is both negative and non-toxic. One approach to such controlled generation is multi-objective reinforcement learning (RL), but how to best combine multiple objectives in a reward function is an open question. In this paper, we investigate various formulations of multi-style rewards, including calibrated outputs from discriminators and dynamic weighting by discriminator gradient magnitudes. We find that our proposed dynamic weighting outperforms static weighting approaches with respect style control while maintaining linguistic quality, and we explore its effectiveness in 2- and 3--style control.
All code and data for the RL pipelines will be publicly available.\footnote{\url{https://github.com/minnesotanlp/dynamic-multi-reward-weighting}}
\end{abstract}

\section{Introduction}
\input{tex/intro}

\section{Related Work}
\input{tex/related}

\section{Proposed Method: Multi-Reward Control in RL}

\input{tex/methods}

\section{Experiment Setup}\label{sec:experiments}

\input{tex/experiments}

\section{Results}

\input{tex/exp_results}

\section{Discussion}
\input{tex/discussion}

\section{Limitations}
\input{tex/limitations}

\section{Ethics Statement}
\input{tex/ethics}

\section*{Acknowledgements}
This work was supported by Sony Research.
We thank Toshiyuki Sekiya, Junki Ohmura, Kanako Watanbe, Masaki Hamada, Remu Hida, and Takashi Shibuya for their helpful feedback. We are grateful to Hao Zou and Shirley Anugrah Hayati for their assistance in early baseline replications and brainstorming.
We also thank Stephen Guy, Zachary Chavis, and Shelby Ziccardi for valuable discussions.

\bibliography{anthology,custom}

\newpage
\appendix

\section{Appendix}
\label{sec:appendix}
\input{tex/appendix}
\end{document}

%% file: tex/intro.tex

Textual style is an important component of communication that conveys information not included in the text's raw semantic content \citep{hovy1995multifunctionality}. 
Consequently, it is vital that language models can understand and apply styles themselves. 
\begin{figure}[t!]
    \centering\vspace{-7mm}
    \includegraphics[trim={1cm 2.5cm 1cm 0cm},clip,width=0.9\columnwidth]{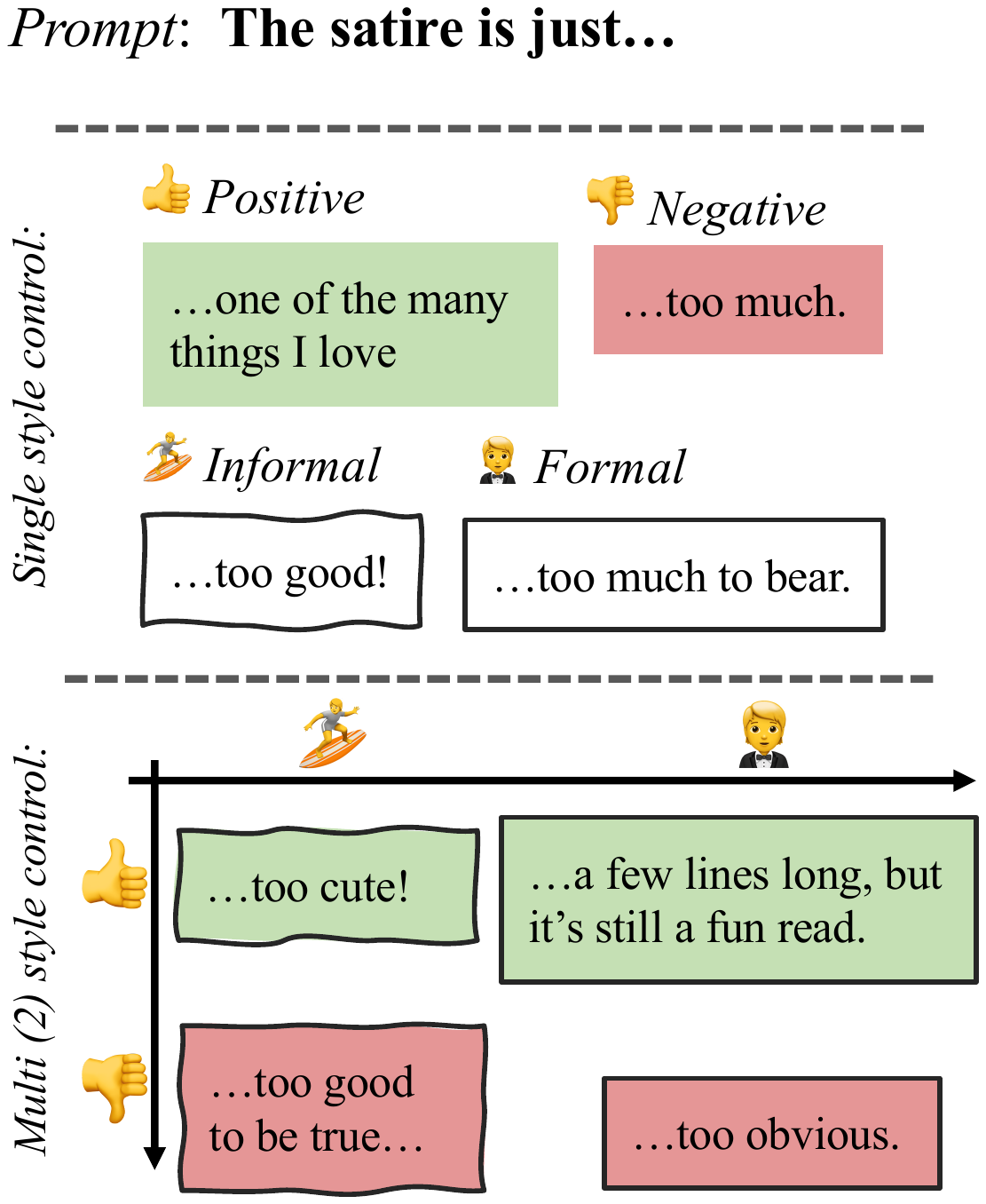}
    \caption{An example of our 1- and 2--style models generating completions to a given prompt. Models are trained with reinforcement learning where the reward is derived from the target styles' discriminators.
    \vspace{-4mm}
    }
    \label{fig:summary}
\end{figure}
Prior work has explored the domain of controlled style generation, a task in which a generative language model aims to generate text with a specified style\footnote{Style transfer, on the other hand, involves paraphrasing given text to a target style without altering its original meaning}. 
However in practice, text frequently contains not only a single style, but a combination of styles \cite{kang2021style}. For example, consider being asked to give feedback to a colleague at work: both formal and positive styles would be appropriate. 
On the other hand, if speaking with a friend about a movie, both informal and positive styles are likely to be useful. Examples of generations following multiple styles models can be found in Figure~\ref{fig:summary}. 

Especially as large language models (LLMs) grow in capability and popularity, it is desirable to include fine-grained control of the styles in LLM outputs. 
For instance, in almost all cases, toxicity and hate speech must be tightly controlled such that the model does not produce harmful output. 
At the same time, in response to the user's preferences or the application, it can be beneficial for the LLM to simultaneously control additional attributes such as humor, formality, or the use of figurative language. 
In order to achieve these goals reliably, techniques for robust multi-style control are needed.\footnote{While some sociolinguistics theories distinguish between textual style and textual attributes, in this work, we follow the common convention in recent NLP papers of broadly using `style' to encompass both of these ideas \cite{jin-etal-2022-deep}.} 

Controlling for more than one style during generation is an under-investigated area, with prior work focusing on controlling for a single style, or a style and a target topic(s) \citep{keskar2019ctrl, liu-etal-2022-multi-attribute}. 
In this work, we investigate the use of Reinforcement learning (RL) for controlling multiple styles. RL approaches satisfy multiple desiderata for generations by employing a reward function in which each individual desideratum contributes to the reward; this approach is recently gaining more interest in the alignment literature (see e.g., fine-grained reinforcement learning from human feedback (RLHF) from \citet{wu2023finegrained}). In this work, we apply a similar approach for multi-style controlled generation, in which ``style scores'' from individual style discriminators are combined into a single reward function during RL fine-tuning. 

The optimal approach to combining multiple reward signals into a reward function is an open question. To further explore this question, we implement several strategies for formulating the multi-style reward, including a novel dynamic-weighting approach. Interestingly, our evaluations indicate that dynamically weighting each component outperforms static weighting, and we also find that simple steps such as confidence calibration and binarization of style discriminator output can improve model performance. We also implement a custom plug-and play pipeline \citep{Dathathri2020Plug} for comparison.

This is to our knowledge a first-of-its-kind work investigating multi-style controllable generation through an RL lens with a new reward shaping approach via dynamic gradient-based weighting. Work on multi-reward formulation for RL is especially relevant in the current landscape, given the modern alignment techniques that incorporate multiple axes of human feedback, as recent work investigates composing multiple types of human feedback -- e.g., helpfulness, correctness -- into the reward function \cite{rame2023rewarded}.

%% file: tex/related.tex


\begin{figure*}[ht]
    \centering
    \includegraphics[trim={0cm 0cm 0cm 0.5cm},clip,width=0.85\textwidth]{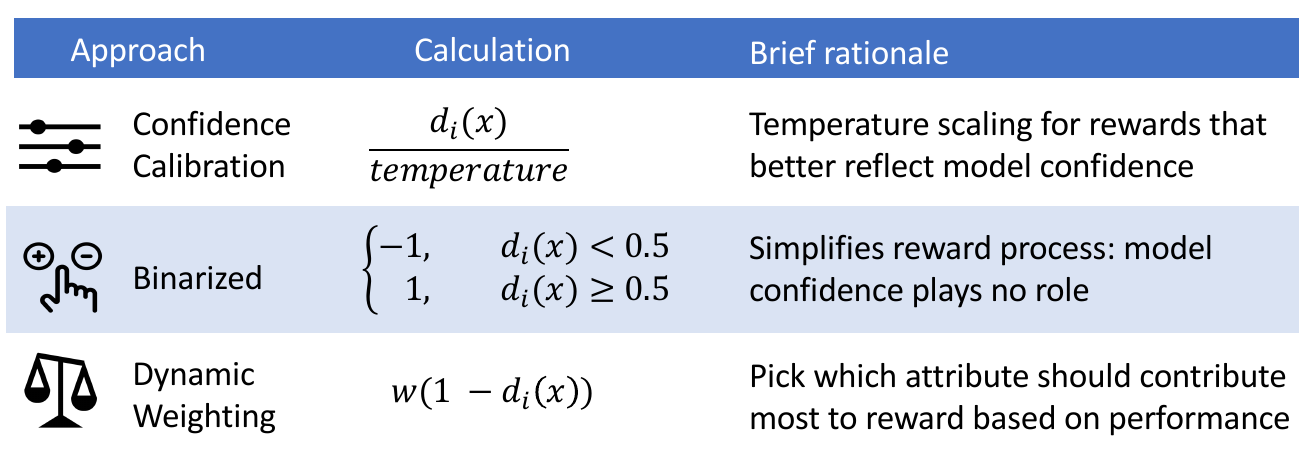}
    \caption{
   Three different approaches to combine multiple rewards effectively for reinforcement learning. 
   $d_i(x)$ refers to the attribute from discriminator $i$ on input text $x$. We find dynamic weighting (our proposal) is particularly effective for multi-style control.}
    \label{fig:enter-label}\vspace{-1em}
\end{figure*}

\paragraph{Controlled Text Generation}
Methods for controlled generation can generally be grouped into three main categories: fine-tuning, retraining, and post-processing \cite{zhang2023survey}. Post-processing approaches are the most lightweight and involve applying transformations during decoding, rather than making any adjustments to model weights themselves. Examples of such methods include plug and play, or \textsc{PPLM} \cite{Dathathri2020Plug}), which uses gradients from an attribute classifier to guide the language model's hidden state; generative discriminators (GeDI) which compute control codes and anti-control codes for all possible next tokens \cite{krause2020gedi}; and Attribute Alignment, which learns an alignment function \cite{yu2021attribute} infuse attribute representations into a pre-trained language model to guide  generation. Prefix-tuning \citep{li-liang-2021-prefix, qian2022controllable} can also guide generation by prepending task- or style-specific ``prefix'' vectors. Retraining (or refactoring) methods involve retraining language models from the ground up on the control task; for example, \textsc{CTRL} \cite{keskar2019ctrl} retrains a class-conditional language model conditioned on many control codes to guide generations. Another retraining approach is Cev-LM \cite{moorjani2024cev}, a prototype-then-edit semi-autoregressive language model that applies edit vectors in the latent space. 

Our work falls under the fine-tuning category. Fine-tuning methods adjust parameters of a pre-trained LLM toward fulfilling the desired controls. Reinforcement learning (RL) is a common fine-tuning approach for controlled text generation \cite{zhang2023survey}, e.g., \citet{gong2019reinforcement} use a style classifier model to provide a target style reward, and \citet{upadhyay2022efficient} use token-level dense rewards and taking the weighted sum of these rewards that were heuristically determined to update the policy. Other works align models towards specific attributes by modeling the reward function as a preference model \cite{rafailov2023direct} to bypass the need for explicitly calculating a reward function, reducing the task to a maximum likelihood objective.  More recently, various approaches to fine-tuning language models via human preferences \cite{ziegler2020finetuning, ouyang-etal-2022-impact} have seen success in guiding text generations to be more aligned with desired attributes. 

Finally, we point out that textual style transfer is related to controlled generation, but it is a distinct task that involves transforming an input text's style while preserving the semantics. Recent work in style transfer include Steering Vectors \citep{subramani-etal-2022-extracting}, which inject steering vectors into the model during decoding. Variations such as ActAdd \cite{turner2023activation} and activation engineering \cite{konen2024style} have also been proposed. 

\paragraph{Multi-Objective Rewards}
Recent work explores how to incorporate reward signals from multiple sources into reward functions, particularly in alignment literature involving RLHF. Notably, recent work investigates multi-objective RLHF by training separate reward models from human preference data and linearly combining those rewards \cite{wu2023finegrained, rame2024rewarded, rame2024warm}. In this work, we apply a similar approach for controlled style generation, and we also experiment with additional approaches to reward combination. 

\paragraph{Inverse RL}
Our work in dynamically shaping the weights of the reward function is also adjacent to Inverse Reinforcement Learning (IRL), which learns both a policy and a reward function. IRL alternates policy update steps with reward model update steps, requiring expert demonstrations for the reward update step. For example, \citet{ghosh-etal-2021-helpful} use IRL for Table-to-Text generation, and their reward function takes several table descriptors as inputs; and \citet{fu-etal-2023-inverse} use IRL for the text summarization task, with a reward function that combines various sub-rewards like saliency and coverage. However, we clarify a distinction between our work and IRL: whereas IRL explicitly learns a reward function, our dynamic weighting method is best characterized as reward shaping for online policy update. Specifically, we do not use expert demonstrations to optimize the weights of the reward function, but instead dynamically "shape" the coefficients of the fine-grained reward function components during the policy update step by inspecting the gradients for each component.

%% file: tex/methods.tex
In this section we provide a preliminary formulation of using reinforcement learning (RL) for a single style control and then formulate the specific multi-reward control functions we propose.

\begin{figure*}[htbp]
     \centering
     \includegraphics[trim={2cm 9cm 5cm 4cm},clip,width=0.8\textwidth]{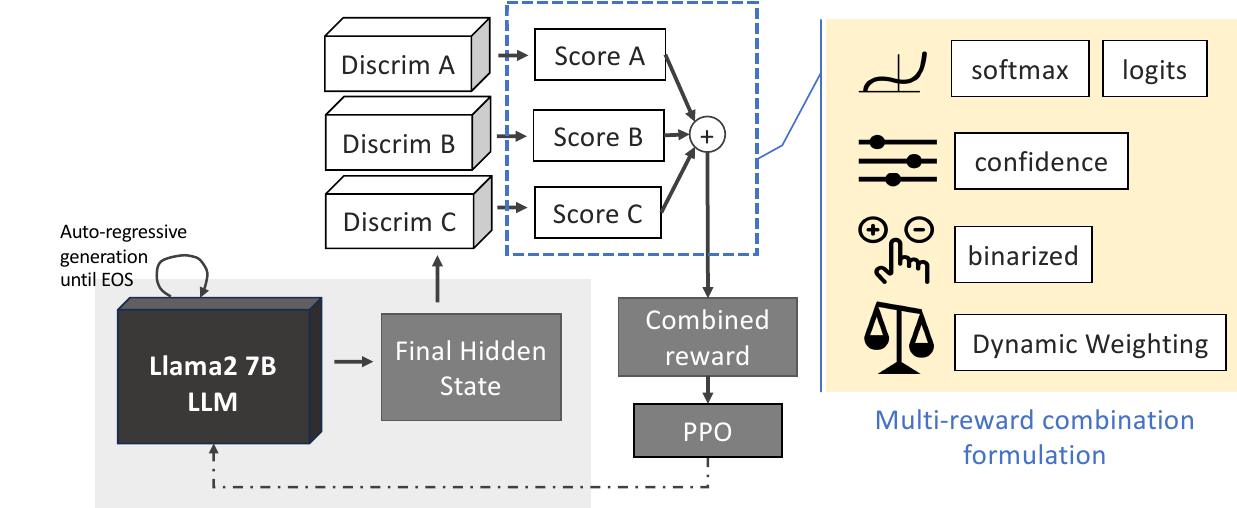}
     \caption{Pipeline for three-style control using fine-tuning Llama 7B model with Proximal Policy Optimization (PPO). We investigate several techniques for integrating feedback from style discriminators into the reward.\label{fig:pipeline}}
\end{figure*}

\subsection{Preliminary Formulation: Style Control using Reinforcement Learning}\label{sec:pre}
Reinforcement learning for language models frames the generative language model as a \textit{policy} network, $\pi_\theta$. The policy is a probability distribution over all possible actions (i.e., vocabulary tokens) that determines $a_t$, i.e., the action to be taken at timestep $t$ given the state $x_{t - 1}$. 
We can then generate tokens based on $\pi_\theta(a_t | x_{t - 1})$. Reinforcement learning also introduces a reward function $R$ that takes in a state $x_t$ and outputs a scalar valuation of that state. (Reinforcement learning for language models generally uses sparse rewards, i.e., a reward is only computed once the full sequence of tokens have been generated.) The objective of reinforcement learning, then, is to learn a new policy $\pi_{\theta'}$ such that the policy maximizes the expected value of $R$. Controlling a single style using RL fine-tuning typically formulates the reward using a discriminator $d$ for the target style, as in $R(x) = d(x)$.


\subsection{Multi-Reward Control Formulations}\label{sec:reward_formulations}
When constructing a reward formulation that combines outputs from multiple style discriminators, we consider two important aspects. The first is the discriminator output itself: we want to find a way to express the output, e.g., via some transformation, that provides a strong and consistent reward signal. 

The second aspect we consider is the how to weigh the combined reward. This is important because training can become intractable if the signal from one discriminator dominates. For instance, consider a 2-style softmax reward formulation, and imagine our starting policy is such that there is a near-zero likelihood of producing a generation that results in high softmax scores for both $d_1$ and $d_2$, but there is a high likelihood of producing generations that results in a high score for $d_1$. The policy will quickly discover how to move in a direction that maximizes scores from $d_1$ and move away from exploring the states that result in more balanced rewards. Instead, we want to combine the discriminators in a way that encourages balancing all outputs.



Motivated by these considerations, we explore multiple approaches to calculate a reward $R$ for a generation $x$ by combining output from the attribute discriminators $d_1, d_2, ..., d_n$ with corresponding target styles $k_1, k_2, ..., k_n$, and we write the logit value for the target class as $d_i(x)_{k_i}$. We consider the following multi-reward formulations:

\paragraph{Logits} We take the logits of the target class from the discriminator output, $R = \sum_{i = 1}^n d_i(x)_{k_i}$.

\paragraph{Softmax} We take the softmax $\sigma$ values of the target class from the discriminator output, $R = \sum_{i = 1}^n \sigma(d_i(x))_{k_i}$.

\paragraph{Binarized} When discriminators are unsure of their prediction, the reward signal becomes noisier, potentially hampering the policy learning process. 
Thus, we consider binarized rewards to make the signals more discrete and emphasized.
\begin{equation}
    R = \sum_{i = 1}^n w_i, \quad w_i = \begin{cases}
        1 & \sigma(d_i(x))_{k_i} \geq 0.5 \\
        -1 & \text{ otherwise }
    \end{cases}
\end{equation}
\paragraph{Calibrated} Logit scores for are adjusted following \citet{guo2017calibration} in order to calibrate model confidence. Similar to the binarized reward, the calibrated reward helps better calibrate the low confidence in the discriminators' predictions to find a more accurate policy. 
\begin{equation}
    R = \sum_{i = 1}^n \frac{d_i(x)_{k_i}}{\texttt{temperature}_{i}}
\end{equation}

\begin{table*}[t]
\centering
\small
\begin{tabular}{l|p{5em}p{5em}p{4em}p{12em}p{4em}}
\hline
& \textbf{Sentiment}         
& \textbf{Formality}        
& \textbf{Irony}                 
& \textbf{Emotion (Eckman Seven)}               
& \textbf{Toxicity}              
\\ \hline
Labels   
& positive, \hspace{3em}negative & formal,\hspace{3em}informal 
& ironic,\hspace{3em}not ironic    
& Fear, anger, joy, sadness, disgust, surprise, neutral                     & toxic, \hspace{3em}non-toxic      
\\\hline
Dataset & SST2 & GYAFC & SemEval & go-emotions & Jigsaw \\\hline
F1 (Macro) & 0.90              & 0.90             & 0.78                  & 0.53                       & 0.80       \\\hline
\end{tabular}
\caption{We train discriminators on each of the above styles (columns). 
}
\label{table:discrims}
\end{table*}

\paragraph{Dynamic Weighting} We calculate a dynamic weight $w_i$. The motivation for the dynamic weighting approach is that when combining multiple style discriminators, it is not always clear whether a signal from one discriminator should be prioritized over another. Our approach weighs the result from each discriminator by considering the magnitude of the gradient of the cross entropy loss for $d_i(x)$ with respect to the desired style. Then, our reward function is
\begin{equation}
    R = \sum_{i = 1}^n w(1 - \sigma(d_i(x))_{k_i})
\end{equation}
\vspace{-3mm}
\begin{equation}
    w = \begin{cases} 
        \mathtt{grad\_norm}_i & \text{ if } \sigma(d_i(x))_{k_i} > 0.5 \\
        -\mathtt{grad\_norm}_i & \text{ otherwise }
    \end{cases}
\end{equation}
where $\mathtt{grad\_norm}_i$ is the normalized magnitude of $d_i(x)$'s gradient:
\begin{equation}
    \texttt{grad\_norm}_i = \frac{||\nabla_{d_i(x)} \mathcal{L}_{CE}||}{\sum_{j = 0}^n ||\nabla_{d_j(x)} \mathcal{L}_{CE}||}
\end{equation} 
We compose all reward shaping methods by taking the convex combination of each discriminator output, with only the gradient scaling method differing. For example, the binarized, confidence, and calibration methods are all follow the formulation with given $d_1$ and $d_2$ (where each $d_n$ refers to some style classifier) 
\begin{equation}
    R_{\text{final}} = \alpha \cdot d_1(x) + (1-\alpha) \cdot d_2(x)
\end{equation}
where each $d_1$ and $d_2$ rewards are shaped via each respective method. Here we take the $\alpha$ as $1 / n$ where $n$ is the number of discriminators and can be extended to any number $n$ such that $\sum_{i=1}^n \alpha_i = 1$. For the gradient-weight scaling, we replace $\alpha$ with \texttt{grad\_norm}$_i$ (extendable to any $n$ number of discriminators) 
\begin{equation}
   R_{\text{final}} =  \texttt{grad\_norm}_1 \cdot d_1(x) + \texttt{grad\_norm}_2 \cdot d_2(x)
\end{equation}

%% file: tex/experiments.tex
This section describes the details of our experimental setup, including baseline models, discriminator training, RL training, and evaluation set.

\subsection{Base Models}
We use LLaMA2 7B \cite{touvron2023llama} as the base model for both the discriminators and the RL pipeline as shown in Figure~\ref{fig:pipeline}. We train discriminators for sentiment, formality, irony, emotion, and toxicity using the SST2, GYAFC, SemEval-Irony, Go-emotions, and Jigsaw Toxicity datasets \cite{socher-etal-2013-recursive, rao2018dear, van2018semeval, demszky2020goemotions, jigsaw}. We train these custom discriminators rather than using existing classifiers because classifiers with the same base model architecture are needed for the PPLM \cite{Dathathri2020Plug} pipeline. For better comparison with PPLM, we use these same custom discriminators in the RL fine-tuning as well. The discriminators are evaluated on the test sets with macro F1 and achieve results comparable to those published in the original dataset papers (Table~\ref{table:discrims}). 

\subsection{Configurations of RL Fine-Tuning}
Figure \ref{fig:pipeline} illustrates our pipeline for RL fine-tuning. The language model generates a completion given a prompt from our dataset. The pooled final hidden state from the model is input to each target style discriminator, whose outputs are then combined into a reward. We use the TRL library \cite{von-werra-etal-2022-evaluate}
implementation of the proximal policy optimization (PPO) algorithm \cite{schulman2017proximal}. Due to computational constraints, we use low rank adapters (LoRA, \citet{hu2022lora}) implemented by the parameter efficient fine-tuning (PEFT, \citet{peft}) library.

The PPO objective includes a penalty term for the Kullback–Leibler (KL) divergence between the fine-tuned and original language model. During training, we use an adaptive KL control setting and do a parameter search for the initial KL coefficient in the range of $[0.2, 0.4]$, eliminating any runs that result in final KL divergences over 20. (Larger KL divergence values in our experience were associated with repetitive reward hacking behaviors.)

\subsection{Configurations of (Custom) PPLM }
We re-implement the \textsc{PPLM} code in order to use it with a \textsc{LLaMAv2} base model rather than GPT-2, as in the original. Our implementation also extends \textsc{PPLM} to consider feedback from multiple discriminators defined by taking a backward step in the hidden state along the gradient for all of the attribute discriminators, $d_1, ..., d_n$. Specifically, the gradient is for the loss function defined by 
\begin{equation}
    \nabla_{d(x)} \mathcal{L} =  \lambda KL(p(x) || p(x')) + \sum_{i = 1}^{n}\ell_{CE}(d_i(x), k_i)
\end{equation}
where $\ell_{CE}$ is the cross entropy loss with respect to the target attribute.

\begin{table}[ht]
\centering
\small
\begin{tabular}{p{0.37\columnwidth}p{0.48\columnwidth}}
\textbf{Source Dataset} & \textbf{Prompt}  \\ \hline
SemEval 2017 & Its nice to stay \\ \hline
TweetEval & @user aha but white \\ \hline
Rotten Tomatoes & One of the most \\ \hline
Wikipedia & Gorman Park (or Amelia \\ \hline
\end{tabular}
\caption{Example prompts from each of the four source datasets used in our pipeline.}
\label{tab:prompts}
\end{table}
\input{Tables/ablation}

\subsection{Training and Evaluation Prompt Data}

Prompts we use during training are drawn from the training sets of these datasets: SemEval emotions \citep{rosenthal2017semeval}, TweetEval  \citep{barbieri2020tweeteval}, Rotten Tomatoes \citep{Pang+Lee:05a}, and Wikimedia \citep{wikidump}.
We choose these datasets for their variety of domains (Tweets, movie reviews, Wikipedia articles). Our evaluation prompts consist of 500 randomly selected items each datasets' test set. Prompts consist of the first four words of each item; examples in Table~\ref{tab:prompts}.

\subsection{Evaluation}
We evaluate the generations from each model based on two criteria: first, how often the generations adhere to the target styles, and second, how well the generations maintain linguistic quality of the original model. Consider a model fine-tuned to produce \textit{positive} generations: it could simply respond to every prompt with ``This is great, I love it!'' and force every generation to be classified as positive, but the overall language quality would suffer.

\noindent \textbf{Automatic Evaluation}
We count the proportion of the generations that are classified by the discriminators as having the target style. We also count the proportion of the generations with \textit{both} target styles to see how frequently models successfully combine styles in the same generation. To evaluate whether the generations maintain the linguistic quality of the original model, we measure the average perplexity of the generations as well as their repetitiveness (duplicate bigram frequency).

\noindent \textbf{Human Evaluation}
To avoid over-reliance on automatic metrics, we also incorporate a human study. Due to financial constraints, the study is on a randomly selected subset of 100 of the evaluation prompts. For each prompt, we collect human preferences between completions from two models. Annotators chose (a) which completion better fulfills the target styles, and (b) which completion sounds most natural. Three Master-qualified annotators from Amazon Mechanical Turk annotate each item, with compensation of $15 \text{ USD}/\text{hour}$. 

%% file: Tables/ablation.tex
\begin{table*}[ht]
\centering
\small
\begin{tabular}{l|ccc|cc}
\toprule
 \textbf{} & \multicolumn{3}{c}{\textbf{Style Accuracy}} & \multicolumn{2}{c}{\textbf{Generation Quality}} 
\\ \midrule
\textbf{Reward Formulation} & \textbf{Negative $\uparrow$} & \textbf{Informal $\uparrow$} & \textbf{Neg \& Inf $\uparrow$} & \textbf{PPL $\downarrow$} & \textbf{Bigram Dup $\downarrow$} \\
\midrule
\quad Softmax & 45.55 & 59.30 & 38.50 & 76.63 & 0.2795 \\
\quad Cal. Softmax & 45.85 & 66.05 & 19.04 & 73.48 & 0.2970  \\
\quad Logits & 56.30 & 74.45 & 52.65 & 98.86 & \textbf{0.1648}  \\
\quad Binary & 62.00 & 76.00 & 56.8 & 32.34 & 0.2800  \\
\quad Dynamic Weighting & \textbf{65.90} & \textbf{76.70} & \textbf{60.25} & \textbf{31.46} & 0.1665 \\
\bottomrule
\end{tabular}
    \caption{Comparison of reward formulations for the \textbf{Negative + Informal} style combinations. We find that our approach (Dynamic Weighting) rewards show the best control over the style combination while maintaining generation quality.
    }
    \label{tab:ablation}
\end{table*}

%% file: tex/exp_results.tex
In this section we describe our experiments for assessing the different multi-style reward formulations and the overall performance of our models for various multi-style control settings. The questions we investigate are: 1. How can we most effectively combine signals from separate style discriminators into a reward function? (\S~\ref{sec:reward_eval}), 2. How often do two- and three-style models express the target styles in their generation, and how fluent are these generations? (\S~\ref{sec:two_style},~\ref{sec:three_styles}), and 3. Which style combinations are most difficult to learn? (\S~\ref{sec:combinations}). 


\input{Tables/compare_base_rl_pplm}
\input{Tables/human_eval}


\subsection{Multi-Reward Formulation Evaluation}\label{sec:reward_eval}
We train several versions of models with Informal-Negative and Formal-Negative target styles to investigate the efficacy of the reward formulations outlined in Sec~\ref{sec:reward_formulations}. For a summary of our study on Informal-Negative see Table~\ref{tab:ablation}; remaining results are in Appendix Table~\ref{tab:ablation-large}. We find that for softmax rewards, the models have minimal style control. We hypothesize that a contributing factor may be poor confidence calibration of the discriminators' softmax scores. To address this, we implement a calibrated softmax reward using the calibration technique in \citet{guo2017calibration}. This results in the Expected Calibration Error (ECE) for sentiment and formality decreasing from $0.133$ to $0.016$ and $0.267$ to $0.111$ respectively. These calibrated softmax rewards do offer some improvements relative to pre-calibration, but improvements are modest.

The logit reward models, on the other hand, tend to have better style control at the cost of fluency -- their generations have some of the highest perplexities. During KL hyperparamater search, we also anecdotally observed that the logit models often developed highly repetitive reward hacking strategies for sub-optimal KL coeffecients.

Both the binary rewards and Dynamic Weighting rewards produce sizable improvements over other reward formulations while maintaining generation quality. Dynamic Weighting has a slight edge overall: it outperforms the binary approach on all counts for the Negative-Informal combination ($60.25\%$ of Dynamic Weighting generations contain both styles versus $56.80\%$ for binary), and its generations are less repetitive. Negative-Formal combination results are more mixed, with a $10.15\%$ difference in \textit{formality} in favor of binary and a $7.56\%$ difference in \textit{negativity} in favor of Dynamic Weighting.
Based on these results, we chose to use the Dynamic Weighting approach for our remaining experiments, since it displays the highest overall performance on both control and generation quality.

Particularly when combining multiple non-orthogonal styles, a simple linear combination of scores can make learning difficult: the model may be able to easily increase its reward by maximizing results from only one discriminator and get ``stuck'' without learning how to obtain more well-rounded results. We conjecture that both the binarizing and dynamic weighting approaches alleviate this issue.

\subsubsection{Human Evaluation}
To bolster the automatic evaluation, we also conduct a human study. For cost reasons, we limit this investigation to two style combinations (negative-informal and negative-formal), and two reward formulations (Logit and Dynamic Weighting). We ask annotators whether they prefer the Logit or Dynamic Weighting completion with respect to style (randomizing the order in which the generations are displayed), and we then ask which completion they prefer with respect to the naturalness of the text. Three annotators respond to each question.

Results indicate that humans prefer the Dynamic Weighting model's generations with respect to both style and linguistic quality (results in Table~\ref{tab:human_eval}), echoing the conclusions of the automatic evaluations. We conclude that the automatic metrics appear to reasonably align with human perception of these aspects of the generation.

We calculated the inter-annotator agreement using Krippendorff’s alpha \cite{krippendorff2018content}. The alpha for style preference questions is 0.36 whereas for linguistic quality questions, alpha is 0.23.This indicates fair agreement among annotators, and we note that it falls in the range of alphas for subjective annotation tasks reported in \citet{wong-etal-2021-cross}.

\subsection{Combinations of Two Styles}\label{sec:two_style}
We first fine-tune RL models for all possible combinations of sentiment and formality: positive+formal, positive+informal, negative+formal, and negative+informal. We choose these style dimensions since their respective discriminators have the highest F1 scores (Table~\ref{table:discrims}).  We use the Dynamic Weighting reward formulation, as it was the most successful reward formulation from Section~\ref{sec:reward_eval}. We consider the generations of the resulting four models on the evaluation prompts: results for these two-style models are in Table~\ref{tab:precision_base-rl-pplm}, and a random sample of their generations are in Appendix Table~\ref{tab:generations}. 

\begin{table}[]
    \centering
    \small
    \begin{tabular}{p{0.29\columnwidth}|p{0.129\columnwidth}p{0.135\columnwidth}p{0.13\columnwidth}p{0.05\columnwidth}}
         Styles &  \% Sent. &  \% Form. & \% I/N/T & PPL \\\hline
         \textbf{Irony}, Pos, For&  85.45 &  77.75 & 66.55 & 39.01 \\\hline
         \textbf{Neutral}, Pos, For& 68.65  & 56.20 & 46.65 & 29.19 \\\hline
         \textbf{Toxic}, Neg, Inf & 57.75 & 67.90  & 18.65 & 29.75\\\hline
         \textbf{Neutral}, Neg, Inf & 55.55  & 73.70 & 52.65 & 39.99
    \end{tabular}
    \caption{We add third style dimension and train a new set of three-style models. Table shows accuracy for each target style dimension (\textbf{Sent}iment, \textbf{Form}ality, and third dimension \textbf{I}rony/\textbf{N}eutral (emotionally)/\textbf{T}oxic) across the 2000 generations in the evaluation set.}
    \label{tab:three-style}
\end{table}


\begin{figure}
    \centering
    \includegraphics[trim={0 0 2cm 0},clip,width=\columnwidth]{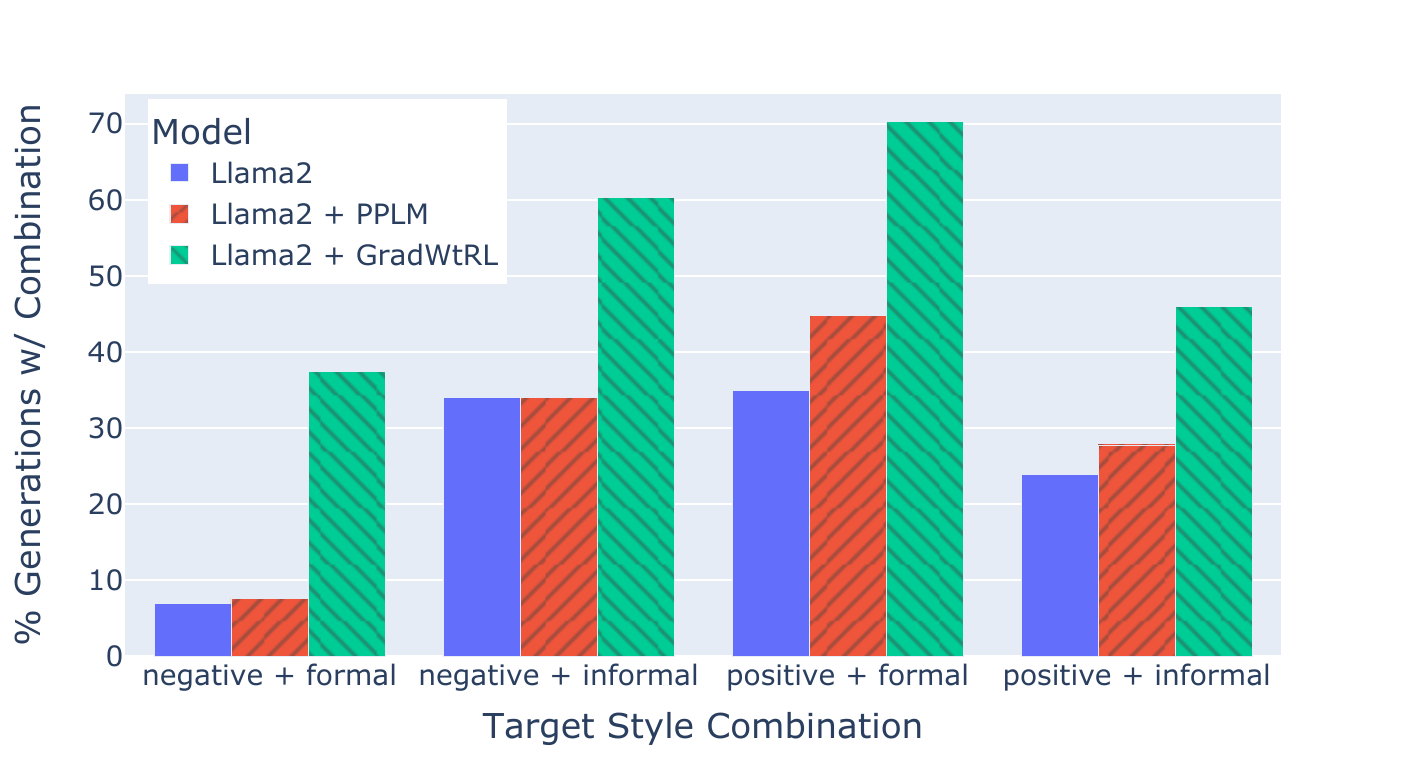}
    \caption{Percentages of evaluation generations that fulfill \textit{both} target styles. Fine-tuning with Dynamic Weighting (GradWt-RL) substantially increases prevalence of target style combinations when compared to the original base model (Llama2) or PPLM-based control.}
    \label{fig:before-and-after}
\end{figure}

\noindent\textbf{Comparison with PPLM}
We extend PPLM to combine feedback from multiple discriminators (Figure ~\ref{fig:before-and-after}) and find that the fine-tuned RL models have higher style control. PPLM has some success with producing more Positive+Formal and Positive+Informal generations than base Llama2, but does not show any improvements for the other combinations. This suggests that PPLM may struggle to effectively combine multiple styles without incorporating some additional techniques. We leave further investigation to future work.

\subsection{Combinations of Three Styles}\label{sec:three_styles}
We further extend the reward formulation to control for three target styles to assess how our approach scales as the number of target styles increases. For three style experiments, we keep the two style dimensions explored earlier (sentiment and formality) and add additional third styles. This decision is motivated in part by a desire to understand to what extent adding a third style control influences the ability to control these dimensions relative to only two styles. We also consider that the sentiment and formality discriminators have the highest F1 scores, making them best-suited to training and evaluation.

\begin{figure}
    \centering 
    \includegraphics[trim={1.2cm 3cm 10cm 0.8cm},clip,width=0.95\columnwidth]{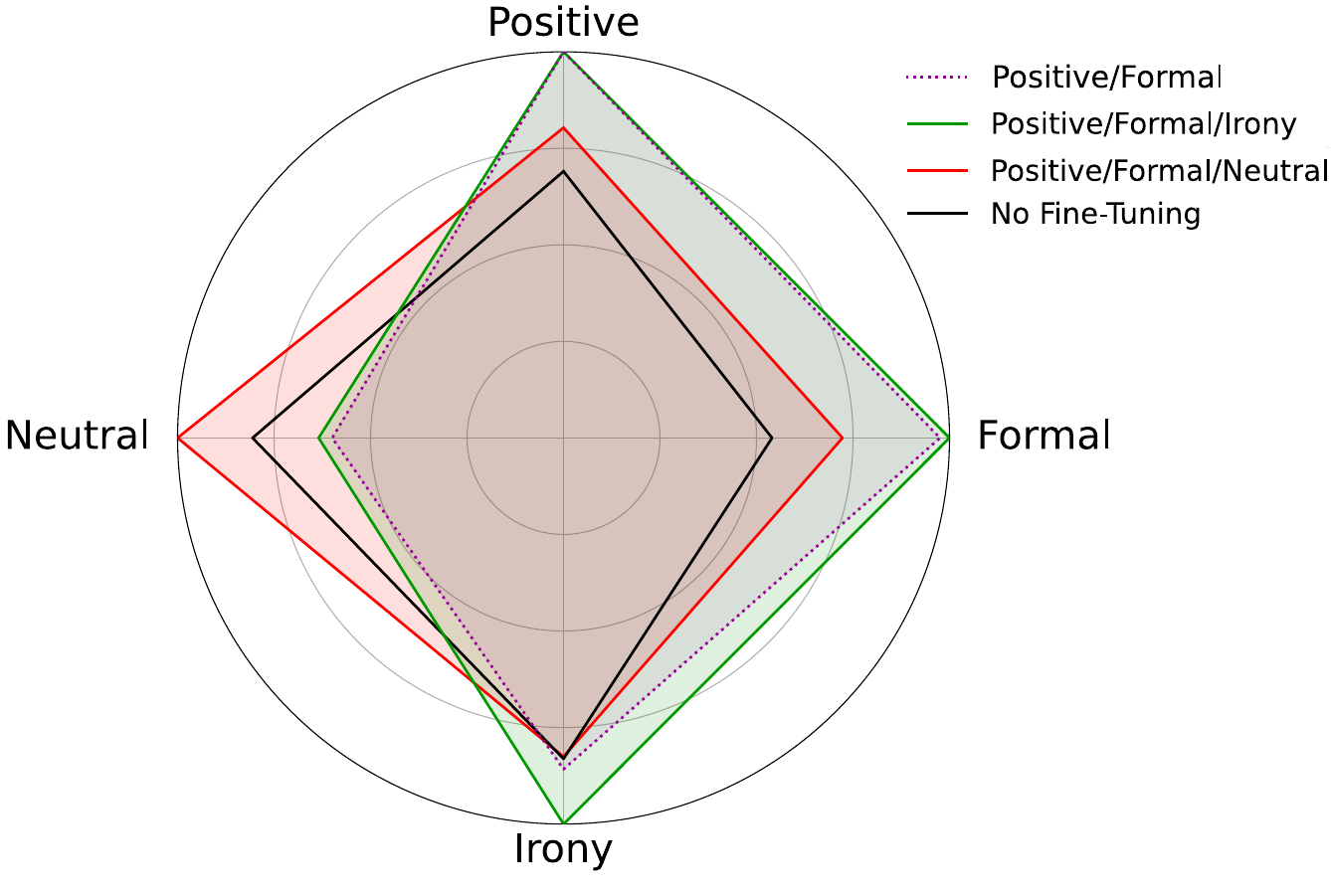}
    \caption{Style control results for two of the 3-style models: Positive+Formal+\textbf{Neutral} (emotion) and Positive+Formal+\textbf{Irony}. Each style dimension shows the portion of generations containing that style and is min-max scaled. The two-style Positive+Formal model results are included for comparison, e.g., we see that the +Neutral model does not control formality as well as the two-style model, but the +Irony model does.}
    \label{fig:three-style}
\end{figure}

The third style dimensions are from our other discriminators: toxicity, irony, and emotion.  We limit our three style models to four distinct three-style combinations due to computational constraints (fully exploring the three-style combination space provided by all of our discriminators would require fine-tuning $212$ models, where each fine-tuning procedure needs to load the language model coupled with three discriminators).
See Fig.~\ref{fig:three-style} for a visualization of these results and Table~\ref{tab:three-style} for a summary. We find that the three style models are able to increase the proportion of all three target styles relative to the base Llama2 model, and their generation quality does not decline. For some combinations (e.g., Positive-Formal-Irony), the sentiment and formality dimension control does not deteriorate compared to the two-style model. Other combinations (e.g., Negative-Informal-Toxic) show lower sentiment and formality control than the two-style models, although they still make improvements over the vanilla model. 

\subsection{Imbalance in Multi-Style Combinations}\label{sec:combinations}
\begin{figure}
    \centering
    \includegraphics[trim={0 0 2cm 0},clip,width=\columnwidth]{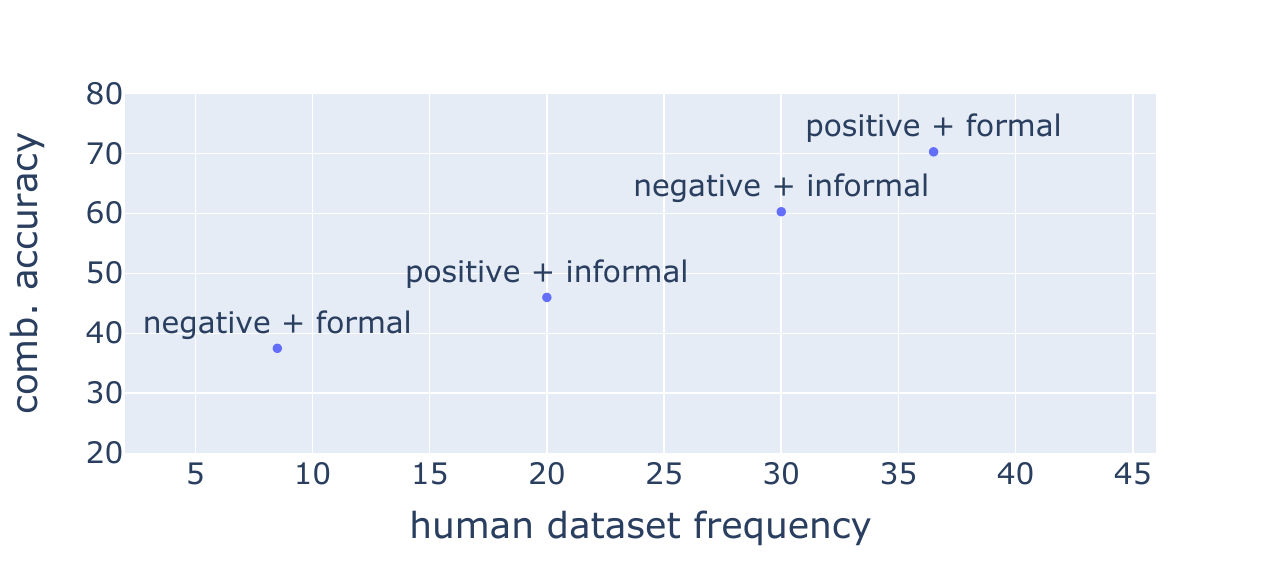}
    \caption{We observe a linear relationship between the percentage of test generations containing both target styles and the prevalence of that target style combination in the full human texts. This suggests that more common style combinations may be easier to learn to control. 
    }
    \label{fig:freq}
    \vspace{-1em}
\end{figure}

We hypothesize that rarer style combinations are more difficult to learn.
\citet{das2023balancing} addressed this style imbalance problem by balancing their combinatory distributions from the training data.
Similarly, we approach understanding the relative frequency of the possible style combinations from two perspectives: frequency in the evaluation datasets' baseline generations and original human texts. In both cases, formal and negative is the least common combination, while formal and positive is most common. Because the training process includes a KL divergence penalty term to maintain linguistic fluency, rare style combinations that are not well-represented in the original model will be suppressed in the controlled model's outputs; Fig.~\ref{fig:freq} illustrates this relationship. We note that our control models are still able to improve frequencies of even the most difficult combinations (Fig.~\ref{fig:before-and-after}).

\subsection{Domain-Based Performance}
When considering the prompts from each of the subdatasets in the evaluation set separately, we observe that the Wikipedia portion of the evaluation set has a substantial drop in performance for Negative-Formal and Positive-Informal models with respect to target style accuracy (Fig.~\ref{sec:appendix}.\ref{fig:domain-results}). 
This indicates that model performance can be strongly affected by the domain of the prompt. It may also indicate that the models struggle to generalize the target style across all domains.

%% file: Tables/compare_base_rl_pplm.tex
\begin{table*}[t!]
\centering
\small
\begin{tabular}{l|ccc|ccc|ccc}
\toprule
\textbf{Model} & \multicolumn{3}{c}{\textbf{Base (Llama2-7B)}} & \multicolumn{3}{c}{\textbf{Dynamic Weighting (ours) }} & \multicolumn{3}{c}{\textbf{PPLM}}
\\ \midrule
Target Styles $\downarrow$& \textbf{Sent.} & \textbf{Form.} & \textbf{Both} & \textbf{Sent.} & \textbf{Form.} & \textbf{Both} & \textbf{Sent.} & \textbf{Form.} & \textbf{Both} \\
\midrule
\quad Positive-Formal & 0.589 & 0.420 & 0.350 & \textbf{0.855} & \textbf{0.759} & \textbf{0.703} & 0.588 & 0.518 & 0.448 \\
\quad Positive-Informal & 0.589 & 0.580 & 0.239 & \textbf{0.750} & \textbf{0.670} & \textbf{0.460} & 0.580 & 0.642  & 0.278 \\
\quad Negative-Formal & 0.411 & 0.420 & 0.070 & \textbf{0.696} & \textbf{0.606} & \textbf{0.375} & 0.432 & 0.528 & 0.076  \\
\quad Negative-Informal & 0.411 & 0.581 & 0.341 & \textbf{0.659} & \textbf{0.767} & \textbf{0.603} & 0.436 & 0.472 & 0.340\\
\bottomrule
\end{tabular}
\caption{Automatic target style accuracy evaluation of 1- and 2-style models for RL and PPLM. (``Both'' indicates the portion of generations containing both the target sentiment \textit{and} formality.) 
All models (Base, Dynamic Weighting, PPLM) use Llama2-7B as the base model for fair comparison.
Compare style frequencies to those in the base LLaMA2 generations to see the effect of the other two approaches.}

\label{tab:precision_base-rl-pplm}
\end{table*}

%% file: Tables/human_eval.tex
\begin{table}[t]
\centering
\small
\begin{tabular}{l|c@{\hskip 1mm}c@{\hskip 1mm}|cc@{\hskip 1mm}}
\toprule
\textbf{Target Styles} & \multicolumn{2}{c}{\textbf{Inf-Neg}}  &\multicolumn{2}{c}{\textbf{Form-Neg}} 
\\ \midrule
& \textbf{Style} & \textbf{Lang} & \textbf{Style} & \textbf{Lang} \\
\midrule
Prefer Logits & $8\%$ & $9\%$ & $8\%$ & $32\%$ \\
Prefer Dynamic Weighting (ours) & \textbf{20}\% & \textbf{69}\% & \textbf{71}\% & \textbf{42}\%\\
\bottomrule
\end{tabular}
\caption{Human evaluation results (preferences were determined by majority vote across annotators). Generations from models trained with our dynamic weighting approach were preferred with respect to both style and linguistic naturalness. (Annotators could also choose no preference.)}

\label{tab:human_eval}
\end{table}

%% file: tex/discussion.tex
We propose an approach to controlled multi-style generation using reinforcement learning with a reward derived from a dynamically weighted linear combination of discriminator outputs. This novel technique results in generations that largely conform to the target styles while maintaining linguistic quality.

There are multiple possible approaches to the multi-style generation problem; our approach represents only one of these. In addition to PPLM, which could possibly be further optimized beyond our additions for multi-style control, there are also the other postprocessing and retraining methods discussed in Section 2, along with other fine-tuning approaches (including Direct Preference Optimization (DPO) \cite{rafailov2023direct}). In addition, this problem could be approached via prompt engineering for LLMs such as GPT-4. Which approach is best-suited to this problem with respect to accuracy, cost, and efficiency remains an open question. Future work should investigate the strongest overall approach multi-style control and the relative accuracy-efficiency trade-offs.

Finally, the precise theoretical relationships between individual styles is an interesting and open question. If two styles rarely co-occur, is it because their combination is impossible, or is it simply rare? Does this distinction affect a language model's ability to combine the two styles? Our results for the three-style models hint at the complexity of this issue. Future work should consider formalizing the feasibility of different low-frequency style combinations, and encouraging language models to explore their state space more in order to uncover more rare combinations of styles.

%% file: tex/limitations.tex
Our focus in this work is mostly centered on combining specific style dimensions, sentiment and formality, and our three-style combinations still include these specific dimensions. We make the decision to focus on specific styles due to computational limits making wider exploration infeasible, and we focus on these particular dimensions due to the high F1 scores of the discriminators for these style dimensions. However, this approach can be extended given more computing resources to further style combinations (e.g. emotional valence and arousal, honesty). It can also be extended reinforcement learning with multiple types of human feedback, as recent work investigates composing multiple types of human feedback -- e.g. helpfulness, correctness -- into the reward function \cite{wu2023finegrained, rame2024rewarded}. 

Our approach is less effective for rare style combinations like negative and formal. For control of rare style combinations, further techniques such as initial language model fine-tuning or exploration incentives are likely needed in order to achieve the desired effect. Our work also relies on discriminators which may be difficult to implement for data scarce attributes.

\paragraph{Alignment Tax} We also note two side effects we observed in the generations after fine-tuning. This is somewhat expected, as many researchers have documented performance on some benchmarking tasks decreasing as side effects of RL fine-tuning procedures (see e.g. \citet{ouyang-etal-2022-impact, askell2021general}). First, there is a possibility that non-target styles may also shift as a result of the fine-tuning process. Specifically, we observe some variations in uncontrolled styles after fine-tuning (e.g. there is more \textit{joy} in positive-formal than in positive-informal; there is more \textit{anger} in negative-formal than in negative-informal. These results are summarized in Appendix Figure~\ref{fig:uncontrolled}. Second, we observe that controlling for style can alter the model's factual claims. The 500-item Wikipedia subset of the evaluation set demonstrates this clearly, as the prompts (e.g. ``Skydio is an American...'') lead the model to write a completion that makes factual assertions. See example Wikipedia generations and our annotations in Appendix A.4 Table~\ref{tab:wiki_generations}. This indicates that further work is needed before such models can be relied on for factual claims.

%% file: tex/ethics.tex
This training approach includes no explicit steering of factuality or truthfulness, which should always be accounted for before deploying LLMs for non-academic use. Further, it is important to be aware of the possibility that controlling for styles such as negativity or disgust can correspondingly lead to increases in toxicity and offensive speech for model generations. 

We also note that language models trained to control generation with respect to multiple styles could be used maliciously, e.g., to create specific voices for counterfeit personas. 

%% file: tex/appendix.tex
\subsection{Further Inspection of 2-Style Models}
\begin{figure}[h]
    \centering
    \includegraphics[trim={0 0 2cm 0},clip,width=\columnwidth]{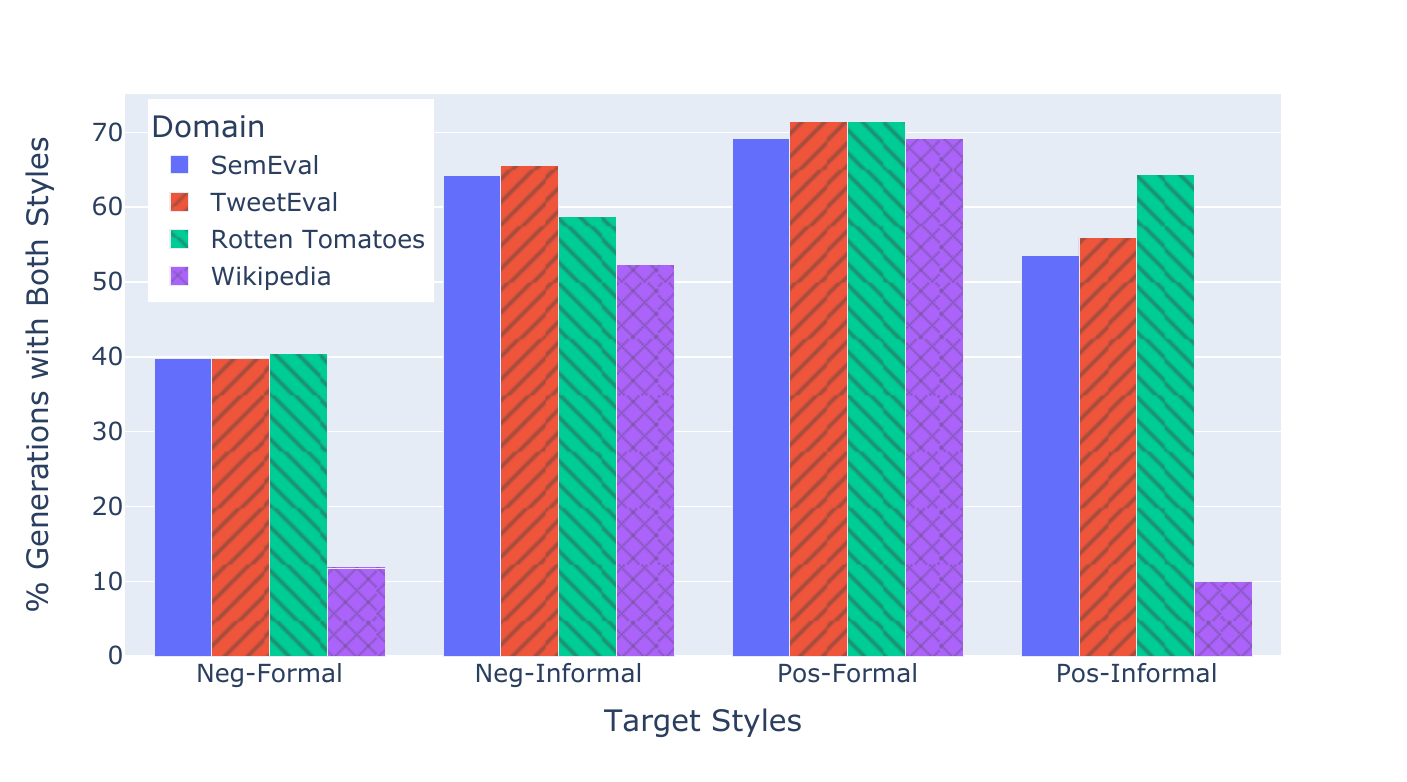}
    \caption{The domain of the four word prompt affects style accuracy.}
    \label{fig:domain-results}
\end{figure}

\begin{figure}[h]
    \centering
    \includegraphics[trim={0 0 2cm 0},clip,width=\columnwidth]{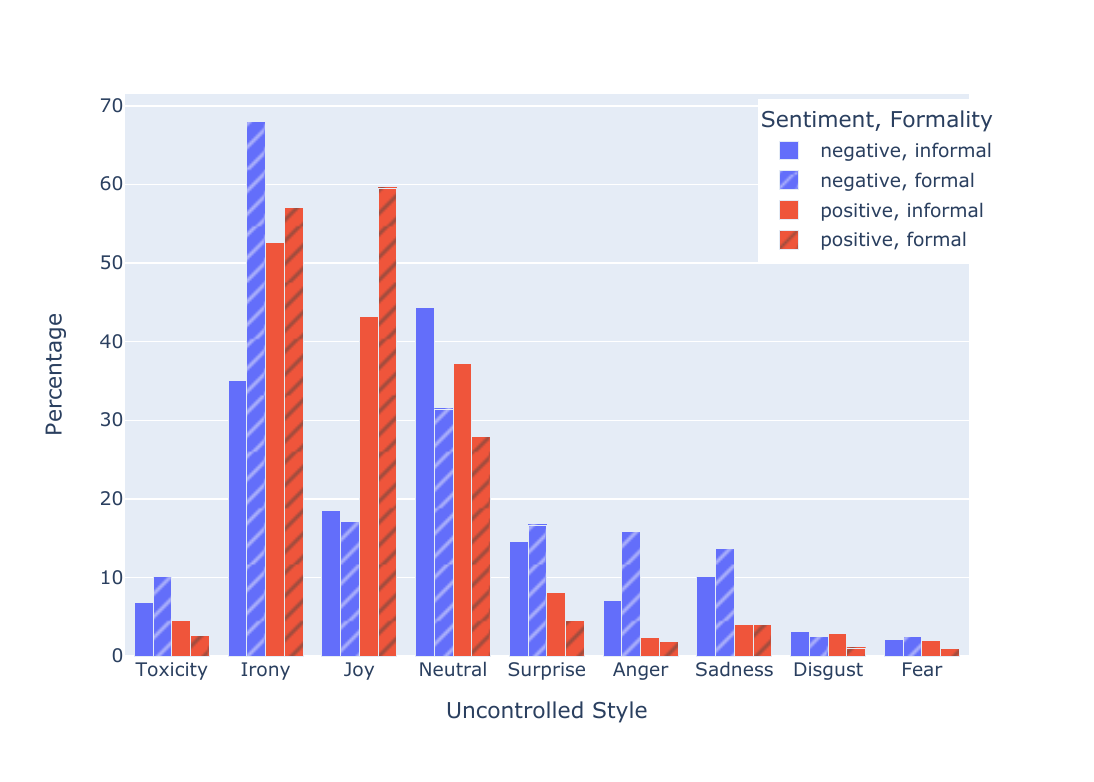}
    \caption{Variations in uncontrolled styles across models.}
    \label{fig:uncontrolled}
\end{figure}

We include graphs of the domain variation in style accuracy across models in Figure~\ref{fig:domain-results}. A visualization of the behavior of uncontrolled attributes is in Figure~\ref{fig:uncontrolled}.

\input{Tables/full_ablation}

\subsection{Additional Reward Formulation Experiments}
We include results for both Informal-Negative and Formal-Negative style combinations in Table~\ref{tab:ablation-large}.

\subsection{Human Study Details}
\begin{figure*}
    \centering
    \includegraphics[width=\textwidth]{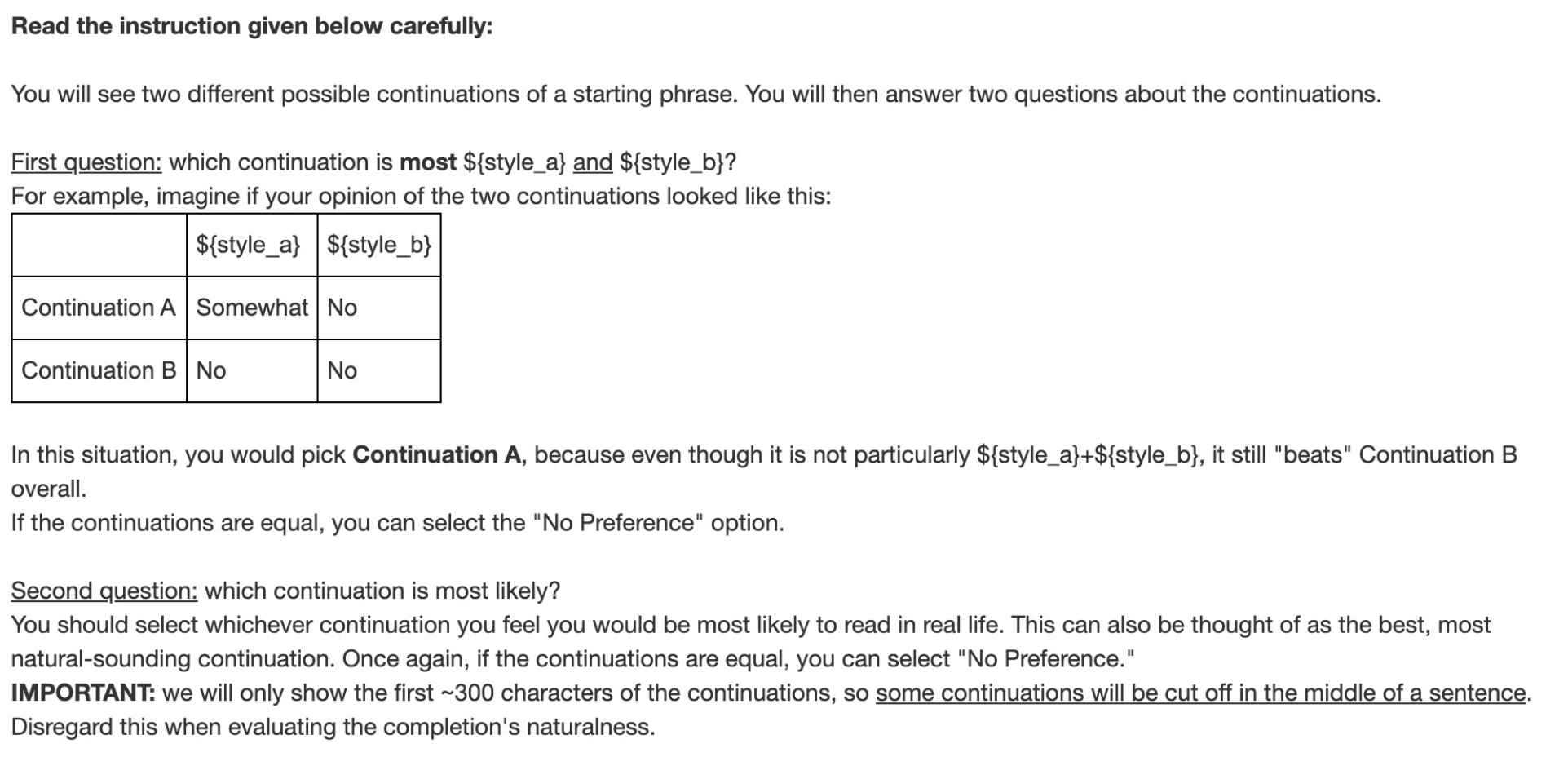}
    \caption{Instructions shown to participants on Amazon Mechanical Turk.}
    \label{fig:amt-inst}
\end{figure*}

The instructions provided to participants are shown in Fig.~\ref{fig:amt-inst}.

\subsection{Factuality Annotations}
We examined a random subset of 60 items from the Wikipedia-based generations, as these tend to encourage the models to make factual claims during generations. We perform this investigation due to concern about factuality shifts after training; for examples, see Table~\ref{tab:wiki_generations}. We hand-annotate the items as shown in Table~\ref{tab:wiki_generations}, and we also check whether the base model generation is factual. We find that for 30\% of the items, the base model's claim is factual. Of these claims, the factual claim was changed about 11.7\% of the time across all fine-tuned models. On the other hand, for 35\% of the claims, the base model's claim is not factual (e.g. the generation ``Alan Thomas (born 7 October 1951) is a former British actor'' -- no such actor exists). In these cases, the fine-tuned models altered claims 65\% of the time. In conclusion, fine-tuned models do have shifted perceptions of facts, and this shift is exagerated in cases where the base model's original claim is incorrect. For 31.2\% of the prompts, no factual claim is made. 

\subsection{GPT-4 Comparison}
\input{Tables/compare_rl_gpt4}
We focus on a basic zero-shot prompting approach to probe GPT-4 performance on controlled style generations. We chose zero-shot prompting as we are interested in scenarios with high data-scarcity, i.e., those in which the target style combination text is not readily available. We note that while few-shot prompting can out-perform zero-shot; this is not always the case -- recent work suggests that performance can be equivalent or even deteriorate in a few-shot setting, e.g. \citet{coyne2023analyzing}.

GPT-4 responses (Table~\ref{tab:precision_rl_gpt4}) indicate that in some scenarios, GPT-4 does very well at combining multiple styles to complete the generation. However, some style combinations (particularly those with an informal style) have markedly worse performance, likely in part due to differences between how our discriminator and GPT-4 understand informality.

We include GPT-4 results as a point of reference, but it is not a direct comparison for a few reasons. First, our other experiments use the Llama2-7B architecture, which has approximately 1,000 times fewer parameters than GPT-4. In addition, Llama2 is a base language model, whereas GPT-4 includes human feedback fine-tuning. Finally, we also recognize that prompting techniques such as Chain-of-Thought prompting \citep{wei2022chain} have been shown to improve performance on many tasks and that prompts can have large impacts on GPT-4 performance; however we do not perform prompt engineering for output optimization.

\subsection{Three-Style Experiment Results}
Numerical results for the three-style control experiments are in Table~\ref{tab:full-three-style}. Example generations can be found in Table~\ref{tab:posformirony}.

\begin{table}[]
\centering
\begin{tabular}{p{0.9\columnwidth}}
\hline
@Ellpeck I fear that might be the case! Thanks for putting up with my nonsense. \\\hline
\#irritated @SouthwestAir I'd be irate too. Thank you for your assistance! \\\hline
Finding your passion seems like an impossible task. But it’s really quite simple! Imagine what you would do if money was no object. \\
\hline
\end{tabular}
\caption{Generations after training for target styles formal, positive, and irony.}\label{tab:posformirony}
\end{table}

\begin{table}[]
    \centering
    \small
    \begin{tabular}{p{0.2\columnwidth}|p{0.15\columnwidth}p{0.15\columnwidth}p{0.13\columnwidth}p{0.1\columnwidth}}
         Styles &  \% Sent. &  \% Form. & \% Other & PPL \\\hline
         \textbf{Irony}, Pos, For&  85.45 (85.45) &  77.75 (75.85) & 66.55 (57.10) & 39.01 \\\hline
         \textbf{Neutral}, Pos, For& 68.65 (75.85) & 56.20 (85.45) & 46.65 (27.95) & 29.19 \\\hline
         \textbf{Toxic}, Neg, Inf & 57.75 (65.90) & 67.90 (76.70) & 18.65 (6.80) & 29.75\\\hline
         \textbf{Neutral}, Neg, Inf & 55.55 (65.90) & 73.70 (76.70) & 52.65 (44.40) & 39.99
    \end{tabular}
    \caption{We add third style dimension such as irony and (emotional) neutrality and train a new set of three-style models. Accuracy scores for these third styles are in the ``\% Other'' column. For reference, we also include the style percentages for generations from the corresponding sentiment-formality two-style models (found in parenthesis).}
    \label{tab:full-three-style}
\end{table}
\subsection{Model Generations}
We include a sample of random generations (i.e., not cherry-picked) from each fine-tuned 2-style control model in Table~\ref{tab:generations}. Generations from Wikipedia prompts are of interest for the shifting factual claims reported by each model; see these generations in Table~\ref{tab:wiki_generations}.

\subsection{GPT-4 Generations}

We include the sample prompt that we used to generate style-controlled completions from the OpenAI API. We utilized the default parameters for generations.
\begin{figure}[h]
  \centering
  \small
  \begin{tabular}{p{0.45\textwidth}}
    \texttt{Please complete the below text, adding no more than 20 - 40 additional words. Make sure the completed text has the following styles: positive.}
    \\ \texttt{\{prompt\}}
  \end{tabular}
\end{figure}

\newpage

\input{Tables/generation_examples}
\input{Tables/wiki_generations}
\input{Tables/accuracy}

%% file: Tables/full_ablation.tex
\begin{table*}[ht]
\centering
\small
\begin{tabular}{l|ccc|cc}
\toprule
 \textbf{} & \multicolumn{3}{c}{\textbf{Style Accuracy}} & \multicolumn{2}{c}{\textbf{Generation Quality}} 
\\ \midrule
\textbf{Reward Formulation} & \textbf{Negative $\uparrow$} & \textbf{Informal $\uparrow$} & \textbf{Neg \& Inf $\uparrow$} & \textbf{PPL $\downarrow$} & \textbf{Bigram Dup $\downarrow$} \\
\midrule
 \textbf{NEGATIVE-INFORMAL models}  & & & & & \\
\quad Softmax & 45.55 & 59.30 & 38.50 & 76.63 & 0.2795 \\
\quad Cal. Softmax & 45.85 & 66.05 & 19.04 & 73.48 & 0.2970  \\
\quad Logits & 56.30 & 74.45 & 52.65 & 98.86 & 0.1648  \\
\quad Binary & 62.00 & 76.00 & 56.80 & 32.34 & 0.2800  \\
\quad GradWt Softmax & 65.90 & 76.70 & 60.25 & 31.46 & 0.1665  \\

\midrule
& \textbf{Negative $\uparrow$} & \textbf{Formal $\uparrow$} & \textbf{Neg \& Form $\uparrow$} & \textbf{PPL $\downarrow$} & \textbf{Bigram Dup $\downarrow$} \\
\midrule
 \textbf{NEGATIVE-FORMAL models} & & & & & \\
\quad Softmax & 44.55 & 47.80 & 10.85 & 25.85 & 0.2275 \\
\quad Cal. Softmax & 43.35 & 53.15 & 13.25 & 26.29 & 0.2243  \\
\quad Logits & 45.00 & 59.50 & 16.35 & 62.52 & 0.2567  \\
\quad Binary & 61.90 & 70.75 & 38.90 & 32.34 & 0.2814  \\
\quad GradWt Softmax & 69.55 & 60.60 & 37.50 & 35.00 & 0.3118 \\

\end{tabular}
    \caption{Comparison of reward formulations for the \textbf{Informal + Negative} and \textbf{Formal + Negative} style combinations. Style accuracy is the percentage of the 2000 evaluations generations classified as having the target style(s).}
    \label{tab:ablation-large}
\end{table*}

%% file: Tables/compare_rl_gpt4.tex
\begin{table}[t]
\centering
\small
\begin{tabular}{l|c@{\hskip 1mm}c@{\hskip 1mm}c|c@{\hskip 1mm}c@{\hskip 1mm}c}
\toprule
\textbf{Model} & \multicolumn{3}{c}{\textbf{Dynamic Weighting}}  &\multicolumn{3}{c}{\textbf{GPT-4}} 
\\ \midrule
Target $\downarrow$& \textbf{Sent.} & \textbf{Form.} & \textbf{Both} & \textbf{Sent.} & \textbf{Form.} & \textbf{Both}  \\
\midrule
Pos-For & 0.855 & 0.759 & 0.703 & \textbf{0.968} & \textbf{0.971} & \textbf{0.945} \\
Pos-Inf & 0.750 & \textbf{0.670} & \textbf{0.460} & \textbf{0.945} & 0.412 & 0.369 \\
Neg-For & \textbf{0.696} & 0.606 & 0.375 & 0.618 & \textbf{0.992} & \textbf{0.608} \\
Neg-Inf & 0.659 & \textbf{0.767} & \textbf{0.603} & \textbf{0.929} & 0.188 & 0.178\\
\bottomrule
\end{tabular}
\caption{Automatic target style accuracy evaluation of 2-style models for Dynamic Weighting and GPT-4.}

\label{tab:precision_rl_gpt4}
\end{table}

%% file: Tables/generation_examples.tex
\begin{table*}
\begin{tabular}{p{0.04\textwidth}|p{0.8\textwidth}|p{0.025\textwidth}p{0.025\textwidth}}

\textbf{} & \textbf{Generation} & \textbf{S} & \textbf{F} \\ \hline\midrule
 \multicolumn{4}{p{0.95\textwidth}}{\textcolor{blue}{Feeling blind. New specs coming, but meanwhile...  My new glasses are on their way, but until I get them I'm stuck with this.  The glasses I have are like two pairs of glasses in }} \\ \hline
P, F & Feeling blind. New specs for 2017!  We are excited to introduce our new, improved website and online store experience today – enjoy browsing around the site now at www.eyesightdirect.com & 0.56 & 0.34 \\ \hline
P, I & Feeling blind. New specs!  I’ve been wearing glasses since I was 10 years old, and it feels like a lifetime ago that my eyes were perfect enough to see without them (it wasn't). So when the time came for me get new pres & 0.33 & 0.74 \\ \hline 
N,F & Feeling blind. New specs aren't helping yet either, but I will try them out soon...  Is there a point to this? & 0.61 & 0.70 \\ \hline 
N, I & Feeling blind. New specs needed...  I've been wearing glasses for about 10 years now, and I need to get a new pair of prescription lenses put into my old frames (which are still in good condition). The problem is that the optician & 0.60 & 0.72 \\ \hline\hline
 \multicolumn{4}{p{0.95\textwidth}}{\textcolor{blue}{Happy Birthday beautiful  We are here to celebrate your birthday and your life}}\\ \hline
P, F & Happy Birthday beautiful!  Thank you so much. I appreciate it very,verymuch!! & 0.73 & 0.68 \\ \hline
P, I & Happy Birthday beautiful <3!!!!! & 0.73 & 0.43 \\ \hline
N,F & Happy Birthday beautiful  I don’t care. It means nothing to me at all, okay? & 0.47 & 0.72 \\ \hline
N, I & Happy Birthday beautiful pic.twitter.com/w60L7zXn5H -- Katie Price (@MissKatiePrice) May 21 & 0.73 & 0.70 \\ \hline \hline
 \multicolumn{4}{p{0.95\textwidth}}{\textcolor{blue}{Fuck me on KIK  Naughty woman would like a night of sex in Killeen  First name: Paulienne, Age: 22 yo, City: Killeen (TX)  Hi, my name is}}\\ \hline
P, F & Fuck me on KIK!  Thanks for checking out my profile. I hope you enjoy your stay here at TheChatAvenue . & 0.72 & 0.68 \\ \hline
P, I & Fuck me on KIK!  Kik: kayla.kiss10 & 0.69 & 0.74 \\ \hline
N,F & Fuck me on KIK!  I'm not interested. Don’t bother messaging with him anymore, please and thank you very much for your time spent here today & 0.38 & 0.36 \\ \hline
N, I & Fuck me on KIK  Fat woman looking for sex in Birmingham Alabama, I'm search fat women who want fuk & 0.57 & 0.72 \\ \hline \hline
\multicolumn{4}{p{0.95\textwidth}}{\textcolor{blue}{Just watched \#TheDarkTower \& it was good. Dark, violent, brutal, entertaining, \& Stephen King's writing style was on full display. It might not be for everyone, but I think}} \\ \hline
P, F & Just watched \#TheDarkTower \& it was great!  Thanks for the review, John. & 0.72 & 0.73 \\ \hline
P, I & Just watched \#TheDarkTower \& it was AWESOME!!!!! Saw The Dark Tower last night! & 0.73 & 0.32 \\ \hline
N,F & Just watched \#TheDarkTower \& it was terrible. I don't get the hype at all! & 0.73 & 0.67 \\ \hline
N, I & Just watched \#TheDarkTower \& it's a mess. @MidnightMovie & 0.56 & 0.55 \\ \hline \hline
 \multicolumn{4}{p{0.95\textwidth}}{\textcolor{blue}{Photo: Orchid Tassel Chain  Inspiration: In the garden  I’ve been thinking a lot lately about how I can make my garden more beautiful, and what I can do to attract more wildlife}} \\ \hline
P, F & Photo: Orchid Tassel Chain  Orchids are a part of the largest family, orchidaea. They have around 250 generas and more than thousand species worldwide! & 0.72 & 0.67 \\ \hline
P, I & Photo: Orchid Tassel Chain Necklace  Orchids are one of my favorite flowers. I love their delicate petals and the way they smell! If you’re like me, then chances… & 0.53 & 0.65 \\ \hline
N,F & Photo: Orchid Tassel Chain  I don't get it. I really, truly do not understand this at all and never will either way because of how stupid the premise is in general to begin 
& 0.63 & 0.28 \\ \hline
N, I & Photo: Orchid Tassel Chain Necklace  Orchids are one of my favorite flowers. I love the colors and shapes, but most importantly they remind me to be patient with myself when things don't go as planned or expected... & 0.52 & 0.73 \\ \hline
\end{tabular}
\caption{Five randomly selected generations (i.e., not cherry-picked) from the evaluation prompts. Original LLaMA2 generations (in \textcolor{blue}{blue}) are followed by generations from fine-tuned models with varying style controls. The models' styles are coded by \textbf{P}ositive / \textbf{N}egative and \textbf{I}nformal / \textbf{F}ormal. \textbf{S}entiment and \textbf{F}ormality classifier outputs for the target classes are also included (values >0.5 indicate that target sentiment or formality is achieved). 
}\label{tab:generations}
\end{table*}

%% file: Tables/wiki_generations.tex
\begin{table*}
\begin{tabular}{p{0.08\textwidth}|p{0.85\textwidth}}

\textbf{Model} & \textbf{Generation}  \\\hline
LLaMA2 & Jo Weldon (born 1962), or "\textcolor{blue}{The Queen of Burlesque}," is an American burlesque performer, producer, teacher, costumer, and historian. Born in \textcolor{blue}{Brooklyn}, New York, W                                                                              \\\hline
P, F   & Jo Weldon (born 1962), a native of \textcolor{red}{San Francisco}, California and an African American woman is the owner founder... Heritage Academies                                                                                                          \\\hline
N, F   & Jo Weldon (born 1962), also known as \textcolor{red}{The Queen of Crump}, is an American-based costume designer. She was born in \textcolor{red}{Philadelphia} and raised by her grandmother on the Lower East Side neighborhoods                                                \\\hline
N, I   & Jo Weldon (born 1962), also known as \textcolor{red}{The Queen of Bling}, is an American costume jewelry designer and owner/operator…                                                                                                                           \\\hline
P, I   & Jo Weldon (born 1962), is an American burlesque performer, producer and writer. She has been described as "the godmother of neo-burlesque". Weldon was born in \textcolor{teal}{New York City} to a Jewish family; her                                                               \\\hline
LLaMA2 & Alan Thomas (born 7 October 1951) is a \textcolor{blue}{former Australian rules footballer} who played for the Geelong Football Club in the Victorian Football League (VFL).  A local Geelong player, Thomas made                                                \\\hline
P, F   & Alan Thomas (born 7 October,1953 in Wolverhampton) is a \textcolor{red}{British composer}. He studied at the Royal Academy of Music with Gordon Crosse and went on to teach there himself from September 2004 until December 6th, when he retired due           \\\hline
N, F   & Alan Thomas (born 7 January 1943 in Edinburgh) is a \textcolor{red}{retired Scottish professional footballer} who played as an inside forward. He made more than fifty appearances for the Scotland national team between March, and October of 196             \\\hline
N, I   & Alan Thomas (born 7 August,1958 in Birmingham) is an \textcolor{red}{English former professional footballer} who played as a defender. Thomas began his career with West Ham United before joining Wolverhampton Wanderers for £200k on transfer deadline day   \\\hline
P, I   & Alan Thomas (born 7 April,1960) is a \textcolor{red}{British actor}. He has appeared in many films and television series since the early eighties including The Krays (2015), The Iron Lady (2013)                                                               \\\hline
LLaMA2 & Dwight is a city in \textcolor{blue}{Illinois}. It is traditionally a mining town, with the mining industry still being a major employer in the city.    \\\hline
P, F   & Dwight is a city in \textcolor{teal}{Illinois}. It was founded by James Baird and named after his son, Charles Hickman "Dutch" Bard who died of pneumonia at the age 23 while serving with Company F during World War I from January to November on              \\\hline
N, F   & Dwight is a city in \textcolor{red}{Michigan}. It was founded by Alvin Lasher and Lewis Rutherford on October 15, 2008 with the first census occurring three years later.{[}6{]} As of June                                                                     \\\hline
N, I   & Dwight is a city in \textcolor{red}{Richland County, Iowa}. Dwight was platted and named by the railroad company that built it through there to connect with another line at Rock Rapids; they chose this name because of its similarity (in sound)to "Wright", \\\hline
P, I   & Dwight is a city in \textcolor{red}{Iowa}. It's located at 41°28'39''N, 096°57'26''W (41.477489,-96.95722). According \\\hline

\end{tabular}
\caption{Examples of Wikipedia prompts that demonstrate the factuality shifting after fine-tuning. We highlight some facts produced by the pre-trained LLaMA2 model in \textcolor{blue}{blue}, with contradictory proclamations by fine-tuned models in \textcolor{red}{red} and compatible proclamations in \textcolor{teal}{green}.
}\label{tab:wiki_generations}
\end{table*}

%% file: Tables/accuracy.tex
\newgeometry{margin=1.5in}
\begin{landscape}
    
\begin{table*}[ht]
\centering
\small
\begin{tabular}{lccc|ccc|ccc|ccc}
\toprule
\textbf{} & \multicolumn{3}{c}{\textbf{Llama2}} & \multicolumn{3}{c}{\textbf{Llama2 + Dynamic Weighting}} & \multicolumn{3}{c}{\textbf{Llama2 + PPLM}} & \multicolumn{3}{c}{\textbf{GPT-4}} 
\\ \midrule
\textbf{} & \textbf{Sent.} & \textbf{Form.} & \textbf{Both} & \textbf{Sent.} & \textbf{Form.} & \textbf{Both} & \textbf{Sent.} & \textbf{Form.} & \textbf{Both} & \textbf{Sent.} & \textbf{Form.} & \textbf{Both} \\
\midrule
1-Style Models & & & & & & \\
\quad Positive & 0.589 & - & - & 0.783 & - & - & 0.602 & - & - &  0.977 & - & -\\
\quad Negative & 0.411 & - & - &0.641 & - & - & 0.472 & - & - & 0.726 & - & -\\
\quad Formal & - & 0.420 & - & - & 0.873 & - & - & 0.562 & - & - & 0.969 & - \\
\quad Informal & - &  0.580 & - & - &0.764 & - & - & 0.466 &  - & - & 0.389 & - \\
\midrule
2-Style Models & & & & & &\\
\quad Positive-Formal & 0.589 & 0.420 & 0.350 & 0.855 & 0.759 & 0.703 & 0.588 & 0.518 & 0.448 & 0.968 & 0.971 & 0.945 \\
\quad Positive-Informal & 0.589 & 0.580 & 0.239 & 0.759 & 0.725 & 0.520 & 0.580 & 0.642  & 0.278 & 0.945 & 0.412 & 0.369 \\
\quad Negative-Formal & 0.411 & 0.420 & 0.07 & 0.696 & 0.606 & 0.375 & 0.432 & 0.528 & 0.076 & 0.618 & 0.992 & 0.608 \\
\quad Negative-Informal & 0.411 & 0.581 & 0.341 & 0.659 & 0.767 & 0.603 & 0.436 & 0.472 & 0.34 & 0.929 & 0.188 & 0.178\\
\bottomrule
\end{tabular}
\caption{Automatic accuracy evaluation of 1- and 2-style models for RL and PPLM.}

\label{tab:precision_overall}
\end{table*}

\end{landscape}

%% file: main.bbl
\begin{thebibliography}{47}
\expandafter\ifx\csname natexlab\endcsname\relax\def\natexlab#1{#1}\fi

\bibitem[{Askell et~al.(2021)Askell, Bai, Chen, Drain, Ganguli, Henighan, Jones, Joseph, Mann, DasSarma et~al.}]{askell2021general}
Amanda Askell, Yuntao Bai, Anna Chen, Dawn Drain, Deep Ganguli, Tom Henighan, Andy Jones, Nicholas Joseph, Ben Mann, Nova DasSarma, et~al. 2021.
\newblock A general language assistant as a laboratory for alignment.
\newblock \emph{arXiv preprint arXiv:2112.00861}.

\bibitem[{Barbieri et~al.(2020)Barbieri, Camacho-Collados, Anke, and Neves}]{barbieri2020tweeteval}
Francesco Barbieri, Jose Camacho-Collados, Luis~Espinosa Anke, and Leonardo Neves. 2020.
\newblock Tweeteval: Unified benchmark and comparative evaluation for tweet classification.
\newblock In \emph{Findings of the Association for Computational Linguistics: EMNLP 2020}, pages 1644--1650.

\bibitem[{Coyne et~al.(2023)Coyne, Sakaguchi, Galvan-Sosa, Zock, and Inui}]{coyne2023analyzing}
Steven Coyne, Keisuke Sakaguchi, Diana Galvan-Sosa, Michael Zock, and Kentaro Inui. 2023.
\newblock Analyzing the performance of gpt-3.5 and gpt-4 in grammatical error correction.
\newblock \emph{arXiv preprint arXiv:2303.14342}.

\bibitem[{Das et~al.(2023)Das, Ma, and Kang}]{das2023balancing}
Debarati Das, David Ma, and Dongyeop Kang. 2023.
\newblock \href {http://arxiv.org/abs/2305.15582} {Balancing effect of training dataset distribution of multiple styles for multi-style text transfer}.

\bibitem[{Dathathri et~al.(2020)Dathathri, Madotto, Lan, Hung, Frank, Molino, Yosinski, and Liu}]{Dathathri2020Plug}
Sumanth Dathathri, Andrea Madotto, Janice Lan, Jane Hung, Eric Frank, Piero Molino, Jason Yosinski, and Rosanne Liu. 2020.
\newblock Plug and play language models: A simple approach to controlled text generation.
\newblock In \emph{International Conference on Learning Representations}.

\bibitem[{Demszky et~al.(2020)Demszky, Movshovitz-Attias, Ko, Cowen, Nemade, and Ravi}]{demszky2020goemotions}
Dorottya Demszky, Dana Movshovitz-Attias, Jeongwoo Ko, Alan Cowen, Gaurav Nemade, and Sujith Ravi. 2020.
\newblock {GoEmotions: A Dataset of Fine-Grained Emotions}.
\newblock In \emph{58th Annual Meeting of the Association for Computational Linguistics (ACL)}.

\bibitem[{Fu et~al.(2023)Fu, Xiong, and Dong}]{fu-etal-2023-inverse}
Yu~Fu, Deyi Xiong, and Yue Dong. 2023.
\newblock \href {https://doi.org/10.18653/v1/2023.findings-emnlp.436} {Inverse reinforcement learning for text summarization}.
\newblock In \emph{Findings of the Association for Computational Linguistics: EMNLP 2023}, pages 6559--6570, Singapore. Association for Computational Linguistics.

\bibitem[{Ghosh et~al.(2021)Ghosh, Qi, Chaturvedi, and Srivastava}]{ghosh-etal-2021-helpful}
Sayan Ghosh, Zheng Qi, Snigdha Chaturvedi, and Shashank Srivastava. 2021.
\newblock \href {https://doi.org/10.18653/v1/2021.acl-short.11} {How helpful is inverse reinforcement learning for table-to-text generation?}
\newblock In \emph{Proceedings of the 59th Annual Meeting of the Association for Computational Linguistics and the 11th International Joint Conference on Natural Language Processing (Volume 2: Short Papers)}, pages 71--79, Online. Association for Computational Linguistics.

\bibitem[{Gong et~al.(2019)Gong, Bhat, Wu, Xiong, and mei Hwu}]{gong2019reinforcement}
Hongyu Gong, Suma Bhat, Lingfei Wu, Jinjun Xiong, and Wen mei Hwu. 2019.
\newblock \href {http://arxiv.org/abs/1903.10671} {Reinforcement learning based text style transfer without parallel training corpus}.

\bibitem[{Guo et~al.(2017)Guo, Pleiss, Sun, and Weinberger}]{guo2017calibration}
Chuan Guo, Geoff Pleiss, Yu~Sun, and Kilian~Q Weinberger. 2017.
\newblock On calibration of modern neural networks.
\newblock In \emph{International conference on machine learning}, pages 1321--1330. PMLR.

\bibitem[{Hovy(1995)}]{hovy1995multifunctionality}
Eduard~H Hovy. 1995.
\newblock The multifunctionality of discourse markers.
\newblock In \emph{Proceedings of the Workshop on Discourse Markers, Egmond-aan-Zee, The Netherlands}.

\bibitem[{Hu et~al.(2022)Hu, Shen, Wallis, Allen-Zhu, Li, Wang, Wang, and Chen}]{hu2022lora}
Edward~J Hu, Yelong Shen, Phillip Wallis, Zeyuan Allen-Zhu, Yuanzhi Li, Shean Wang, Lu~Wang, and Weizhu Chen. 2022.
\newblock \href {https://openreview.net/forum?id=nZeVKeeFYf9} {Lo{RA}: Low-rank adaptation of large language models}.
\newblock In \emph{International Conference on Learning Representations}.

\bibitem[{Jigsaw()}]{jigsaw}
Google Jigsaw.
\newblock \href {https://www.kaggle.com/c/jigsaw-toxic-comment-classification-challenge/data} {Jigsaw toxicity dataset}.

\bibitem[{Jin et~al.(2022)Jin, Jin, Hu, Vechtomova, and Mihalcea}]{jin-etal-2022-deep}
Di~Jin, Zhijing Jin, Zhiting Hu, Olga Vechtomova, and Rada Mihalcea. 2022.
\newblock \href {https://doi.org/10.1162/coli_a_00426} {Deep learning for text style transfer: A survey}.
\newblock \emph{Computational Linguistics}, 48(1):155--205.

\bibitem[{Kang and Hovy(2021)}]{kang2021style}
Dongyeop Kang and Eduard Hovy. 2021.
\newblock Style is not a single variable: Case studies for cross-stylistic language understanding.
\newblock In \emph{Proceedings of the 59th Annual Meeting of the Association for Computational Linguistics and the 11th International Joint Conference on Natural Language Processing (Volume 1: Long Papers)}, pages 2376--2387.

\bibitem[{Keskar et~al.(2019)Keskar, McCann, Varshney, Xiong, and Socher}]{keskar2019ctrl}
Nitish~Shirish Keskar, Bryan McCann, Lav~R. Varshney, Caiming Xiong, and Richard Socher. 2019.
\newblock \href {http://arxiv.org/abs/1909.05858} {Ctrl: A conditional transformer language model for controllable generation}.

\bibitem[{Konen et~al.(2024)Konen, Jentzsch, Diallo, Sch{\"u}tt, Bensch, Baff, Opitz, and Hecking}]{konen2024style}
Kai Konen, Sophie Jentzsch, Diaoul{\'e} Diallo, Peer Sch{\"u}tt, Oliver Bensch, Roxanne~El Baff, Dominik Opitz, and Tobias Hecking. 2024.
\newblock Style vectors for steering generative large language model.
\newblock \emph{arXiv preprint arXiv:2402.01618}.

\bibitem[{Krause et~al.(2020)Krause, Gotmare, McCann, Keskar, Joty, Socher, and Rajani}]{krause2020gedi}
Ben Krause, Akhilesh~Deepak Gotmare, Bryan McCann, Nitish~Shirish Keskar, Shafiq Joty, Richard Socher, and Nazneen~Fatema Rajani. 2020.
\newblock \href {http://arxiv.org/abs/2009.06367} {Gedi: Generative discriminator guided sequence generation}.

\bibitem[{Krippendorff(1980)}]{krippendorff2018content}
Klaus Krippendorff. 1980.
\newblock \emph{Content analysis: An introduction to its methodology}.
\newblock Sage publications.

\bibitem[{Li and Liang(2021)}]{li-liang-2021-prefix}
Xiang~Lisa Li and Percy Liang. 2021.
\newblock \href {https://doi.org/10.18653/v1/2021.acl-long.353} {Prefix-tuning: Optimizing continuous prompts for generation}.
\newblock In \emph{Proceedings of the 59th Annual Meeting of the Association for Computational Linguistics and the 11th International Joint Conference on Natural Language Processing (Volume 1: Long Papers)}, pages 4582--4597, Online. Association for Computational Linguistics.

\bibitem[{Liu et~al.(2022)Liu, Li, Guo, Luo, and Wang}]{liu-etal-2022-multi-attribute}
Guisheng Liu, Yi~Li, Yanqing Guo, Xiangyang Luo, and Bo~Wang. 2022.
\newblock \href {https://aclanthology.org/2022.coling-1.516} {Multi-attribute controlled text generation with contrastive-generator and external-discriminator}.
\newblock In \emph{Proceedings of the 29th International Conference on Computational Linguistics}, pages 5904--5913, Gyeongju, Republic of Korea. International Committee on Computational Linguistics.

\bibitem[{Mangrulkar et~al.(2022)Mangrulkar, Gugger, Debut, Belkada, and Paul}]{peft}
Sourab Mangrulkar, Sylvain Gugger, Lysandre Debut, Younes Belkada, and Sayak Paul. 2022.
\newblock Peft: State-of-the-art parameter-efficient fine-tuning methods.
\newblock \url{https://github.com/huggingface/peft}.

\bibitem[{Moorjani et~al.(2024)Moorjani, Krishnan, and Sundaram}]{moorjani2024cev}
Samraj Moorjani, Adit Krishnan, and Hari Sundaram. 2024.
\newblock Cev-lm: Controlled edit vector language model for shaping natural language generations.
\newblock \emph{arXiv preprint arXiv:2402.14290}.

\bibitem[{Ouyang et~al.(2022)Ouyang, Ye, and Li}]{ouyang-etal-2022-impact}
Siqi Ouyang, Rong Ye, and Lei Li. 2022.
\newblock \href {https://doi.org/10.18653/v1/2022.iwslt-1.9} {On the impact of noises in crowd-sourced data for speech translation}.
\newblock In \emph{Proceedings of the 19th International Conference on Spoken Language Translation (IWSLT 2022)}, pages 92--97, Dublin, Ireland (in-person and online). Association for Computational Linguistics.

\bibitem[{Pang and Lee(2005)}]{Pang+Lee:05a}
Bo~Pang and Lillian Lee. 2005.
\newblock Seeing stars: Exploiting class relationships for sentiment categorization with respect to rating scales.
\newblock In \emph{Proceedings of the ACL}.

\bibitem[{Qian et~al.(2022)Qian, Dong, Shen, Wei, and Chen}]{qian2022controllable}
Jing Qian, Li~Dong, Yelong Shen, Furu Wei, and Weizhu Chen. 2022.
\newblock \href {http://arxiv.org/abs/2202.13257} {Controllable natural language generation with contrastive prefixes}.

\bibitem[{Rafailov et~al.(2023)Rafailov, Sharma, Mitchell, Ermon, Manning, and Finn}]{rafailov2023direct}
Rafael Rafailov, Archit Sharma, Eric Mitchell, Stefano Ermon, Christopher~D Manning, and Chelsea Finn. 2023.
\newblock Direct preference optimization: Your language model is secretly a reward model.
\newblock \emph{arXiv preprint arXiv:2305.18290}.

\bibitem[{Rame et~al.(2024)Rame, Couairon, Dancette, Gaya, Shukor, Soulier, and Cord}]{rame2024rewarded}
Alexandre Rame, Guillaume Couairon, Corentin Dancette, Jean-Baptiste Gaya, Mustafa Shukor, Laure Soulier, and Matthieu Cord. 2024.
\newblock Rewarded soups: towards pareto-optimal alignment by interpolating weights fine-tuned on diverse rewards.
\newblock \emph{Advances in Neural Information Processing Systems}, 36.

\bibitem[{Rame et~al.(2023)Rame, Couairon, Shukor, Dancette, Gaya, Soulier, and Cord}]{rame2023rewarded}
Alexandre Rame, Guillaume Couairon, Mustafa Shukor, Corentin Dancette, Jean-Baptiste Gaya, Laure Soulier, and Matthieu Cord. 2023.
\newblock Rewarded soups: towards pareto-optimal alignment by interpolating weights fine-tuned on diverse rewards.
\newblock \emph{arXiv preprint arXiv:2306.04488}.

\bibitem[{Ram{\'e} et~al.(2024)Ram{\'e}, Vieillard, Hussenot, Dadashi, Cideron, Bachem, and Ferret}]{rame2024warm}
Alexandre Ram{\'e}, Nino Vieillard, L{\'e}onard Hussenot, Robert Dadashi, Geoffrey Cideron, Olivier Bachem, and Johan Ferret. 2024.
\newblock Warm: On the benefits of weight averaged reward models.
\newblock \emph{arXiv preprint arXiv:2401.12187}.

\bibitem[{Rao and Tetreault(2018)}]{rao2018dear}
Sudha Rao and Joel Tetreault. 2018.
\newblock Dear sir or madam, may i introduce the gyafc dataset: Corpus, benchmarks and metrics for formality style transfer.
\newblock In \emph{Proceedings of the 2018 Conference of the North American Chapter of the Association for Computational Linguistics: Human Language Technologies, Volume 1 (Long Papers)}, pages 129--140.

\bibitem[{Rosenthal et~al.(2017)Rosenthal, Farra, and Nakov}]{rosenthal2017semeval}
Sara Rosenthal, Noura Farra, and Preslav Nakov. 2017.
\newblock Semeval-2017 task 4: Sentiment analysis in twitter.
\newblock In \emph{Proceedings of the 11th International Workshop on Semantic Evaluation (SemEval-2017)}, pages 502--518.

\bibitem[{Schulman et~al.(2017)Schulman, Wolski, Dhariwal, Radford, and Klimov}]{schulman2017proximal}
John Schulman, Filip Wolski, Prafulla Dhariwal, Alec Radford, and Oleg Klimov. 2017.
\newblock Proximal policy optimization algorithms.
\newblock \emph{arXiv preprint arXiv:1707.06347}.

\bibitem[{Socher et~al.(2013)Socher, Perelygin, Wu, Chuang, Manning, Ng, and Potts}]{socher-etal-2013-recursive}
Richard Socher, Alex Perelygin, Jean Wu, Jason Chuang, Christopher~D. Manning, Andrew Ng, and Christopher Potts. 2013.
\newblock \href {https://aclanthology.org/D13-1170} {Recursive deep models for semantic compositionality over a sentiment treebank}.
\newblock In \emph{Proceedings of the 2013 Conference on Empirical Methods in Natural Language Processing}, pages 1631--1642, Seattle, Washington, USA. Association for Computational Linguistics.

\bibitem[{Subramani et~al.(2022)Subramani, Suresh, and Peters}]{subramani-etal-2022-extracting}
Nishant Subramani, Nivedita Suresh, and Matthew Peters. 2022.
\newblock \href {https://doi.org/10.18653/v1/2022.findings-acl.48} {Extracting latent steering vectors from pretrained language models}.
\newblock In \emph{Findings of the Association for Computational Linguistics: ACL 2022}, pages 566--581, Dublin, Ireland. Association for Computational Linguistics.

\bibitem[{Touvron et~al.(2023)Touvron, Martin, Stone, Albert, Almahairi, Babaei, Bashlykov, Batra, Bhargava, Bhosale, Bikel, Blecher, Ferrer, Chen, Cucurull, Esiobu, Fernandes, Fu, Fu, Fuller, Gao, Goswami, Goyal, Hartshorn, Hosseini, Hou, Inan, Kardas, Kerkez, Khabsa, Kloumann, Korenev, Koura, Lachaux, Lavril, Lee, Liskovich, Lu, Mao, Martinet, Mihaylov, Mishra, Molybog, Nie, Poulton, Reizenstein, Rungta, Saladi, Schelten, Silva, Smith, Subramanian, Tan, Tang, Taylor, Williams, Kuan, Xu, Yan, Zarov, Zhang, Fan, Kambadur, Narang, Rodriguez, Stojnic, Edunov, and Scialom}]{touvron2023llama}
Hugo Touvron, Louis Martin, Kevin Stone, Peter Albert, Amjad Almahairi, Yasmine Babaei, Nikolay Bashlykov, Soumya Batra, Prajjwal Bhargava, Shruti Bhosale, Dan Bikel, Lukas Blecher, Cristian~Canton Ferrer, Moya Chen, Guillem Cucurull, David Esiobu, Jude Fernandes, Jeremy Fu, Wenyin Fu, Brian Fuller, Cynthia Gao, Vedanuj Goswami, Naman Goyal, Anthony Hartshorn, Saghar Hosseini, Rui Hou, Hakan Inan, Marcin Kardas, Viktor Kerkez, Madian Khabsa, Isabel Kloumann, Artem Korenev, Punit~Singh Koura, Marie-Anne Lachaux, Thibaut Lavril, Jenya Lee, Diana Liskovich, Yinghai Lu, Yuning Mao, Xavier Martinet, Todor Mihaylov, Pushkar Mishra, Igor Molybog, Yixin Nie, Andrew Poulton, Jeremy Reizenstein, Rashi Rungta, Kalyan Saladi, Alan Schelten, Ruan Silva, Eric~Michael Smith, Ranjan Subramanian, Xiaoqing~Ellen Tan, Binh Tang, Ross Taylor, Adina Williams, Jian~Xiang Kuan, Puxin Xu, Zheng Yan, Iliyan Zarov, Yuchen Zhang, Angela Fan, Melanie Kambadur, Sharan Narang, Aurelien Rodriguez, Robert Stojnic, Sergey Edunov, and Thomas
  Scialom. 2023.
\newblock \href {http://arxiv.org/abs/2307.09288} {Llama 2: Open foundation and fine-tuned chat models}.

\bibitem[{Turner et~al.(2023)Turner, Thiergart, Udell, Leech, Mini, and MacDiarmid}]{turner2023activation}
Alex Turner, Lisa Thiergart, David Udell, Gavin Leech, Ulisse Mini, and Monte MacDiarmid. 2023.
\newblock Activation addition: Steering language models without optimization.
\newblock \emph{arXiv preprint arXiv:2308.10248}.

\bibitem[{Upadhyay et~al.(2022)Upadhyay, Sudhakar, and Maheswaran}]{upadhyay2022efficient}
Bhargav Upadhyay, Akhilesh Sudhakar, and Arjun Maheswaran. 2022.
\newblock \href {http://arxiv.org/abs/2204.07696} {Efficient reinforcement learning for unsupervised controlled text generation}.

\bibitem[{Van~Hee et~al.(2018)Van~Hee, Lefever, and Hoste}]{van2018semeval}
Cynthia Van~Hee, Els Lefever, and V{\'e}ronique Hoste. 2018.
\newblock Semeval-2018 task 3: Irony detection in english tweets.
\newblock In \emph{Proceedings of The 12th International Workshop on Semantic Evaluation}, pages 39--50.

\bibitem[{Von~Werra et~al.(2022)Von~Werra, Tunstall, Thakur, Luccioni, Thrush, Piktus, Marty, Rajani, Mustar, and Ngo}]{von-werra-etal-2022-evaluate}
Leandro Von~Werra, Lewis Tunstall, Abhishek Thakur, Sasha Luccioni, Tristan Thrush, Aleksandra Piktus, Felix Marty, Nazneen Rajani, Victor Mustar, and Helen Ngo. 2022.
\newblock \href {https://aclanthology.org/2022.emnlp-demos.13} {Evaluate {\&} evaluation on the hub: Better best practices for data and model measurements}.
\newblock In \emph{Proceedings of the 2022 Conference on Empirical Methods in Natural Language Processing: System Demonstrations}, pages 128--136, Abu Dhabi, UAE. Association for Computational Linguistics.

\bibitem[{Wei et~al.(2022)Wei, Wang, Schuurmans, Bosma, Xia, Chi, Le, Zhou et~al.}]{wei2022chain}
Jason Wei, Xuezhi Wang, Dale Schuurmans, Maarten Bosma, Fei Xia, Ed~Chi, Quoc~V Le, Denny Zhou, et~al. 2022.
\newblock Chain-of-thought prompting elicits reasoning in large language models.
\newblock \emph{Advances in Neural Information Processing Systems}, 35:24824--24837.

\bibitem[{Wikimedia()}]{wikidump}
Wikimedia.
\newblock \href {https://dumps.wikimedia.org} {Wikimedia downloads}.

\bibitem[{Wong et~al.(2021)Wong, Paritosh, and Aroyo}]{wong-etal-2021-cross}
Ka~Wong, Praveen Paritosh, and Lora Aroyo. 2021.
\newblock \href {https://doi.org/10.18653/v1/2021.acl-long.548} {Cross-replication reliability - an empirical approach to interpreting inter-rater reliability}.
\newblock In \emph{Proceedings of the 59th Annual Meeting of the Association for Computational Linguistics and the 11th International Joint Conference on Natural Language Processing (Volume 1: Long Papers)}, pages 7053--7065, Online. Association for Computational Linguistics.

\bibitem[{Wu et~al.(2023)Wu, Hu, Shi, Dziri, Suhr, Ammanabrolu, Smith, Ostendorf, and Hajishirzi}]{wu2023finegrained}
Zeqiu Wu, Yushi Hu, Weijia Shi, Nouha Dziri, Alane Suhr, Prithviraj Ammanabrolu, Noah~A. Smith, Mari Ostendorf, and Hannaneh Hajishirzi. 2023.
\newblock \href {http://arxiv.org/abs/2306.01693} {Fine-grained human feedback gives better rewards for language model training}.

\bibitem[{Yu et~al.(2021)Yu, Yu, and Sagae}]{yu2021attribute}
Dian Yu, Zhou Yu, and Kenji Sagae. 2021.
\newblock \href {http://arxiv.org/abs/2103.11070} {Attribute alignment: Controlling text generation from pre-trained language models}.

\bibitem[{Zhang et~al.(2023)Zhang, Song, Li, Zhou, and Song}]{zhang2023survey}
Hanqing Zhang, Haolin Song, Shaoyu Li, Ming Zhou, and Dawei Song. 2023.
\newblock A survey of controllable text generation using transformer-based pre-trained language models.
\newblock \emph{ACM Computing Surveys}, 56(3):1--37.

\bibitem[{Ziegler et~al.(2020)Ziegler, Stiennon, Wu, Brown, Radford, Amodei, Christiano, and Irving}]{ziegler2020finetuning}
Daniel~M. Ziegler, Nisan Stiennon, Jeffrey Wu, Tom~B. Brown, Alec Radford, Dario Amodei, Paul Christiano, and Geoffrey Irving. 2020.
\newblock \href {http://arxiv.org/abs/1909.08593} {Fine-tuning language models from human preferences}.

\end{thebibliography}
